\documentclass{article}

\usepackage[preprint]{neurips_2026}

\usepackage[utf8]{inputenc} 
\usepackage[T1]{fontenc}    
\usepackage{hyperref}       
\usepackage{url}            
\usepackage{booktabs}       
\usepackage{amsfonts}       
\usepackage{nicefrac}       
\usepackage{microtype}      
\usepackage{xcolor}         
\usepackage{multirow}

\usepackage{booktabs}
\usepackage{array}
\usepackage{caption}
\usepackage[T1]{fontenc}
\usepackage{graphicx}
\usepackage{enumitem}
\usepackage{amsmath}
\usepackage{tabularx}
\usepackage{placeins}
\usepackage{makecell}
\usepackage{longtable}

\usepackage{longtable}
\usepackage{adjustbox}
\usepackage{threeparttable}

\newcolumntype{C}[1]{>{\centering\arraybackslash}p{#1}}
\newcolumntype{Y}{>{\raggedright\arraybackslash}X}

\title{Same Compression Principle, Different Geometry: 
Rate--Distortion Signatures Dissociate Biological and Artificial Visual Systems}

\author{%
  \begin{minipage}[t]{0.30\textwidth}
    \centering
    Leyla R. Caglar \\
    {\normalfont
    Windreich Department of AI \\
    and Human Health \\
    Icahn School of Medicine \\
    at Mount Sinai \\
    New York, NY, USA \\}
  \end{minipage}%
  \hspace{0.03\textwidth}%
  \begin{minipage}[t]{0.30\textwidth}
    \centering
    Pedro A.M. Mediano \\
    {\normalfont
    Department of Computing \\
    Imperial College London \\
    London, UK \\}
  \end{minipage}%
  \hspace{0.03\textwidth}%
  \begin{minipage}[t]{0.30\textwidth}
    \centering
    Baihan Lin \\
    {\normalfont
    Windreich Department of AI \\
    and Human Health \\
    Department of Psychiatry \\
    Icahn School of Medicine \\
    at Mount Sinai \\
    New York, NY, USA \\
    \texttt{baihan.lin@mssm.edu}}
  \end{minipage}
}

\begin{document}

\maketitle

\begin{abstract}
Efficient coding theory predicts that biological perceptual systems compress sensory input optimally under resource constraints, with the systematic structure of errors reflecting the geometry of that compression. Here we operationalize this principle using rate--distortion theory (RDT) to characterize how any system - biological or artificial - trades representational fidelity for informational efficiency. Treating stimulus-response behavior as an effective communication channel, we infer rate--distortion (RD) frontiers directly from confusion matrices and summarize each system with three geometric signatures: slope ($\beta$), curvature ($\kappa$), and area under the RD curve (AUC), capturing the marginal cost, abruptness, and overall efficiency of the accuracy-compression trade-off respectively. Applying this framework to human psychophysical data and 18 deep vision models across 12 families of controlled image perturbations at graded severities, we find that both biological and artificial systems follow a common lossy-compression principle but occupy systematically different regions of RD space. Humans exhibit smooth, flexible trade-offs characteristic of near-optimal efficient coding, while deep networks operate in steeper, more brittle regimes even at matched accuracy, with geometry dissociable from performance across training regimes. Critically, behavioral RD signatures track internal representational geometry, evidenced by the behaviorally inferred compression structure correlating with internal representational dissimilarity across all models. Moreover, $\kappa$ and AUC constitute complementary signatures: AUC tracks representational geometry independently of accuracy and captures the efficiency of learned categorical representations, with alignment strengthening specifically at deep categorical layers, while $\kappa$ captures the abruptness of compression transitions at the level of the behavioral RD frontier and shows a distortion-type-dependent relationship with accuracy degradation dynamics. These results establish RD geometry as a compact diagnostic of perceptual compression strategy that recovers mechanistically interpretable structure in internal representations from behavioral input alone and extends naturally to the direct characterization of compression geometry in neural population activity.
\end{abstract}

\section{Introduction}
When you glance at a photograph through a rain-streaked window or see a blurry photograph, your visual system faces an immediate computational problem: the image reaching your retina is degraded, ambiguous, and incomplete and yet you recognize the scene almost effortlessly. This feat of robust perception under noise reflects a fundamental design principle of biological vision. Instead of faithfully transmitting raw sensory information to downstream areas, the visual system compresses it, retaining the structure most useful for behavior while discarding what is redundant or uninformative. This compression is necessarily \textit{lossy} as some fidelity must be sacrificed. Efficient coding theory proposes that biological sensory systems compress optimally under metabolic and channel-capacity constraints \citep{attneave1954informational,barlow1961possible,simoncelli2001natural,atick1992could}, such that the systematic pattern of perceptual errors reflects the structure of that compression, rather than noise. In this view, how a perceptual system \textit{fails} — which stimuli it confuses and under what conditions — is as informative about its computational strategy as how it succeeds.

This principle of efficient coding has proven to be generative across levels of analysis. At the neural level, efficient coding accounts for the tuning properties of retinal ganglion cells, V1 simple cells, and neurons in higher visual areas, predicting the statistics of receptive fields from the statistics of natural images \citep{atick1992could,simoncelli2001natural,olshausen1996emergence}. At the behavioral level, Sims \citep{sims2018efficient,sims2016rate} showed that a foundational regularity in human perception — Shepard's Universal Law of Generalization \citep{shepard1987toward}, which describes how the probability of treating two stimuli as equivalent decays with their dissimilarity in psychological space — follows directly from modeling perception as a system that minimizes information transmission under a fidelity constraint. What appeared to be an empirical regularity specific to human psychology turned out to reflect a universal consequence of optimal compression under resource limits and that the rate at which generalization decays with dissimilarity is set by the information constraints the system operates under, not by any property of stimuli or tasks in particular. While characterizing the rate of generalization decay as a single parameter is powerful, it leaves open how the full geometry of the compression trade-off - its steepness, its nonlinearity, and how it shifts as input quality degrades - is structured across systems and conditions, and whether these geometric properties are grounded in the internal representational structure that implements them. These are precisely the questions we address, exploiting the
complementarity between biological and artificial vision systems, where humans provide the behaviorally rich compression signatures that matter for understanding perception, while artificial networks --- whose internal representations are fully accessible --- provide the ground truth needed to establish whether those signatures genuinely recover the representational geometry that implements compression.

Rate--distortion theory (RDT) \citep{shannon1959coding,cover2006elements} provides the natural formalism for characterizing this trade-off precisely. In lossy compression, not all information can be preserved. A system must sacrifice some representational fidelity in exchange for operating within its information-processing capacity. The rate--distortion function \textit{R(D)} formalizes this, tracing the minimum information rate required to achieve any given level of expected distortion, producing the RDT curve or Pareto frontier between fidelity and efficiency. We propose that the \textit{shape} of this frontier is the key diagnostic of compression strategy (Figure 1a). A system whose frontier has a gentle slope ($\beta$) and low curvature ($\kappa$) compresses gracefully, because as conditions worsen, fidelity declines in a distributed, predictable way. A system with a steep slope and high curvature compresses brittlely, where the system maintains high fidelity until a certain threshold after which performance collapses abruptly. The area under the RD curve (AUC) summarizes overall efficiency across the full range of tolerated distortion. Together, ($\beta$, $\kappa$, $\mathrm{AUC}$) are not merely descriptive summaries but geometric signatures of the compression strategy a system implements that otherwise remain invisible to any single accuracy measurement (Figure 1b).

To apply this framework to biological and artificial vision, we study human observers alongside 18 deep neural network (DNN) models spanning diverse architectures and training regimes. Although DNNs achieve human-level accuracy on standard visual benchmarks \citep{he2016deep,russakovsky2015imagenet}, they fail under image degradation in ways that are qualitatively unlike human perception. Specifically, errors accumulate abruptly rather than gradually, and tend to be concentrated in ways that suggest a fundamentally different compression strategy rather than a noisier version of the same one \citep{geirhos2019imagenet,hendrycks2019benchmarking,drenkow2021systematic}. Characterizing that difference precisely, and grounding it in internal representational structure, requires a framework that operates at the level of the full error geometry rather than its scalar summary. DNNs are an ideal test bed because, unlike biological systems, their internal representations are fully accessible, allowing us to directly validate whether behavioral RD signatures recover the internal representational geometry that implements compression strategy — a validation that is essential before applying these tools to neural population recordings, where behavior can be measured but representations can only be partially observed. We treat each system's stimulus-response behavior as an effective communication channel, infer its rate--distortion frontier from confusion matrices under 12 families of controlled image perturbations at graded severities, and extract $\beta$, $\kappa$, $\mathrm{AUC}$  as system-level signatures (Figure 1c). We find that both humans and DNNs conform to a common lossy-compression principle but occupy systematically different regions of RD space, with humans clustering in a low-$\beta$, low-$\kappa$ regime characteristic of smooth, near-optimal efficient coding while deep networks operate in steeper, more brittle regimes. We further show that training interventions move networks along different axes of the RD surface, revealing side-effects invisible to accuracy-based evaluation. Critically, we demonstrate that behavioral RD signatures are grounded in internal representational geometry: AUC tracks internal representational geometry independently of accuracy, with alignment deepening across the visual processing hierarchy, while $\kappa$ and AUC constitute complementary
signatures of compression strategy at two distinct mechanistic levels. These results establish RD geometry as a compact diagnostic of perceptual compression strategy that recovers mechanistically interpretable structure in internal representations from behavioral data alone — and extends naturally to the direct characterization of compression geometry in neural population activity.

\begin{figure*}[t]
  \centering
  \includegraphics[width=\textwidth]{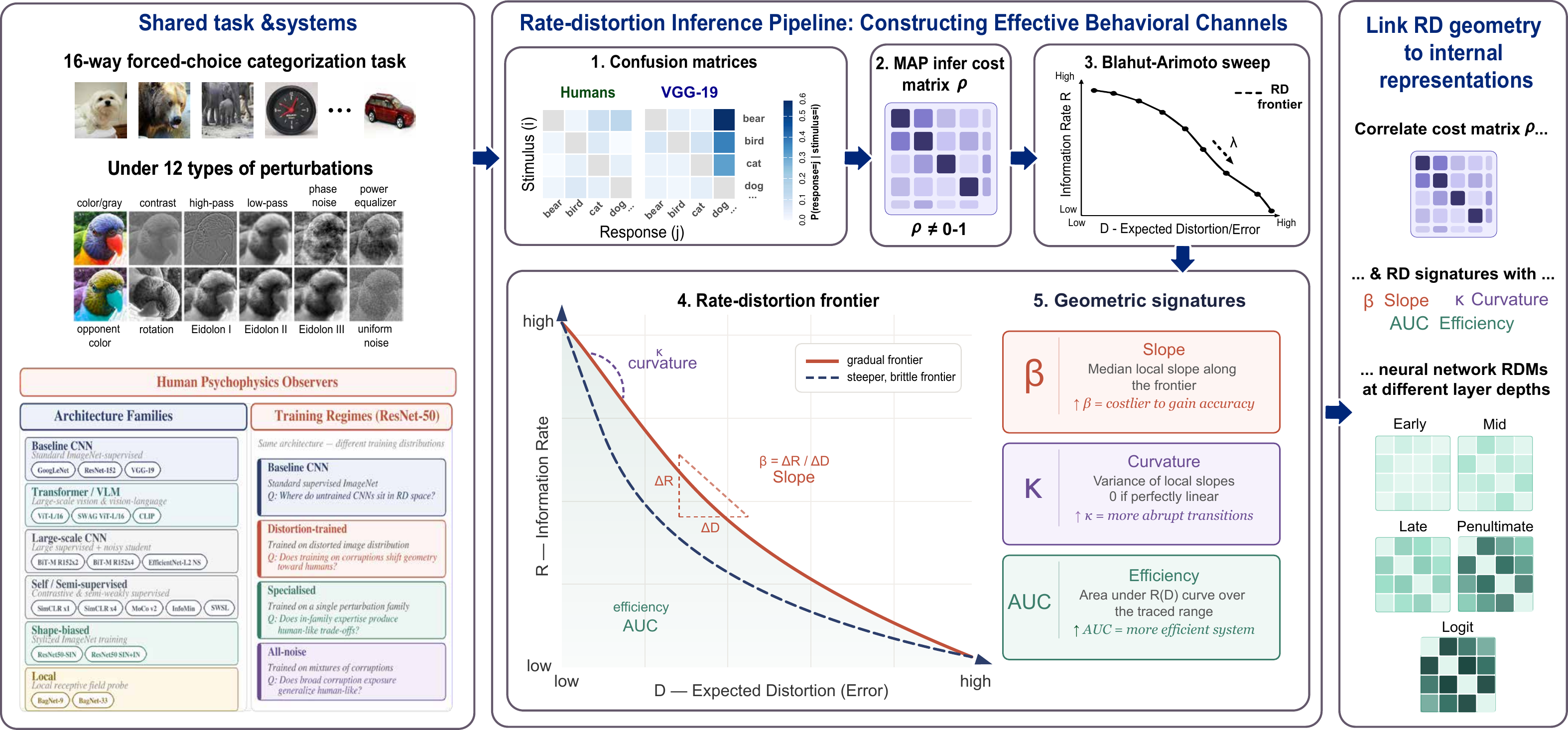}
  \caption{\textbf{Overview of shared task, systems, and analysis pipeline.}
  \textit{Left:} Human psychophysical observers and 18 deep neural network models spanning diverse architecture families and training regimes were evaluated on a shared 16-way forced-choice categorization task under 12 families of controlled
  image perturbations at graded severities (perturbation figure from \citep{geirhos2018generalisation}.
  \textit{Centre:} The rate--distortion inference pipeline proceeds in five steps:
  (1)~stimulus--response behavior is summarized as empirical confusion matrices for each system, perturbation type, and severity level. (2)~a system-specific cost matrix $\rho$ is inferred from the confusion structure via MAP optimization, yielding a graded, non-binary error-cost geometry; (3)~the rate--distortion frontier is traced numerically via Blahut--Arimoto optimization over a grid of inverse-temperature values $\lambda$. (4)~the resulting frontier describes the Pareto trade-off between information rate $R$ and expected distortion $D$, with schematic examples illustrating gradual (low-$\beta$, low-$\kappa$) versus steep, brittle (high-$\beta$, high-$\kappa$) frontiers. (5)~three geometric signatures are extracted from each frontier: slope ($\beta$), curvature ($\kappa$), and efficiency (AUC), quantifying integrated performance over the traced distortion range. \textit{Right:} To validate that behavioral RD signatures recover internal representational structure, the inferred cost matrix $\rho$ and the geometric signatures $(\beta, \kappa, \mathrm{AUC})$ are correlated with representational dissimilarity matrices (RDMs; \citep{kriegeskorte2008representational}) computed from network activations at five layer depths (early through logit).}
\label{fig:schematic}
\end{figure*}

\section*{Results}

To characterize perceptual compression strategy across biological and artificial visual systems, we treat each system's stimulus--response behavior as an effective communication channel estimated from empirical confusion matrices, and infer its rate--distortion frontier directly from the structure of errors under controlled image perturbations. Rather than assuming a fixed error cost, we infer a system-specific distortion geometry $\rho$ from the pattern of confusions via maximum-a-posteriori (MAP) optimization and then trace the corresponding RD frontier numerically across a range of compression demands. The shape of this frontier is summarized by three geometric signatures: slope ($\beta$), quantifying the marginal information cost of increasing fidelity; curvature ($\kappa$), quantifying the abruptness of transitions between compression regimes; and area under the RD curve (AUC), quantifying overall efficiency across the range of tolerated distortion (Figure~\ref{fig:schematic}).

We apply this framework to matched psychophysical data from human observers (~83k trials from the GEN repository ~\citep{geirhos2019imagenet}, ~85k trials from the ModelZoo repository \citep{geirhos2021partial}) and 18 deep neural network models drawn from two publicly available repositories~\citep{geirhos2019imagenet,geirhos2021partial}, spanning four broad categories: supervised CNNs ranging from standard ImageNet-trained baselines (GoogLeNet, ResNet-152, VGG-19) to large-scale pretrained models (BiT) and architecturally specialized local-feature networks (BagNet); transformer and vision--language models (ViT-L, SWAG-ViT, CLIP); self- and semi-supervised models (SimCLR, MoCo v2, InfoMin, SWSL, Noisy Student); and shape-biased models trained on Stylized ImageNet (Figure~\ref{fig:schematic}). In addition to these architecture families, we analyze four training regimes. Three regimes are applied to ResNet-50 variants: a distortion-trained regime in which models are trained on a degraded image distribution, and two robustness-oriented regimes - specialised (single-distortion training) and all-noise (multi-corruption training) - that allow controlled comparison of how training distribution affects compression geometry independently of architecture. The standard supervised ImageNet-trained CCN models GoogLeNet, ResNet-152, and VGG-19 evaluated without any robustness-specific intervention serve as \textit{baseline} regimes. All systems were evaluated on the same 16-way ImageNet-derived categorization task across 12 families of controlled image perturbations at graded severities, including contrast reduction, phase scrambling, uniform noise, blur, and Eidolon-family distortions, among others (Figure~\ref{fig:schematic}; see Methods for full details of all systems, datasets, and estimation procedures).

The following results are organized in five sections. We first validate that behavioral confusion structure conforms to RDT predictions across all systems, then characterize how architecture families and training interventions occupy distinct regions of RD space, before establishing that AUC tracks internal representational geometry independently of accuracy while $\kappa$ captures behavioral compression abruptness -- a dissociation that deepens with processing depth -- , and finally show that $\kappa$ predicts the shape of accuracy degradation trajectories across perturbation severity.

\begin{figure*}[t]
  \centering
  \includegraphics[width=\textwidth]{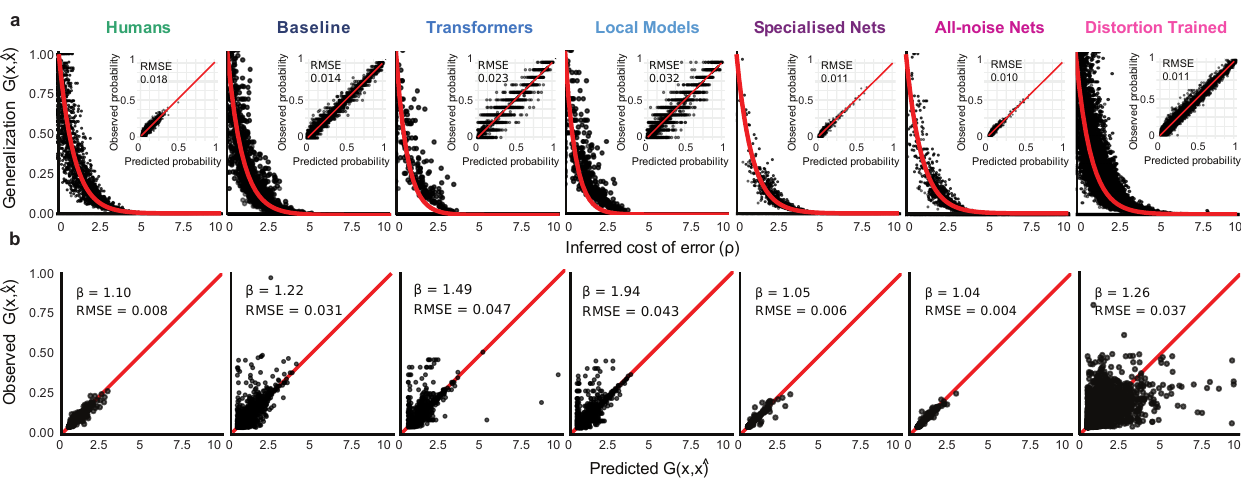}
  \caption{\textbf{Generalization curves and goodness-of-fit of rate--distortion predictions.}
(a) Empirical generalization $G(x,\hat{x})$ (points) plotted against inferred cost of error $\rho$ for humans and selected model families. Red curves show the corresponding exponential/RD-style prediction. Insets show observed vs.\ predicted off-diagonal confusion probabilities (RDT channel prediction), with RMSE reported per family.
(b) Summary of exponential-gradient fits: observed binned generalization gradients versus predictions from the best-fit $a\,e^{-\beta d}$ model (red diagonal indicates equality). Each panel reports the fitted slope $\beta$ and RMSE, quantifying family-specific deviations from a simple exponential generalization law.}
\label{fig:figure1_rdt_fits}
\end{figure*}

\subsection*{Behavioral confusion structure conforms to rate--distortion predictions across biological and artificial systems}

Before characterizing differences in compression geometry across systems, we established that the rate--distortion framework provides a faithful account of behavioral confusion structure for both human observers and all neural network families tested, which is a prerequisite for interpreting geometric signatures as meaningful compression diagnostics rather than fitting artifacts. Across all systems, the inferred RDT channel captures confusion structure with consistently low error (channel-fit RMSE range: 0.010--0.032 across families). Family-specific residuals are themselves informative, showing the tightest fit for All-noise ($\mathrm{RMSE} = 0.010$) and Specialised ($0.011$) models, a good fit for Baseline CNNs ($0.014$) and Humans ($0.018$), and highest for Local Models ($0.032$), reflecting architectural differences in how cleanly confusion structure is captured by a single-channel approximation (Figure~\ref{fig:figure1_rdt_fits}a).

The shape of the generalization gradient encodes how quickly confusion probability decays as a function of inferred error cost and provides a complementary diagnostic. Fitting each system's empirical gradient to the exponential form $ae^{-\beta d}$ yields low errors overall (RMSE range: 0.004--0.047), but with systematic family-level departures that reveal qualitative differences in compression strategy. Human observers and All-noise networks are closest to a pure exponential decay ($\mathrm{RMSE} = 0.008$ and $0.004$ respectively), consistent with smooth, distributed compression. Baseline CNNs show larger gradient-shape deviations ($\mathrm{RMSE} = 0.031$), and Vision Transformers and Local Models show the largest departures ($0.047$ and $0.043$), indicating that these architectures implement qualitatively different and more nonlinear generalization gradients that a single exponential cannot fully capture (Figure~\ref{fig:figure1_rdt_fits}b). That both humans and artificial systems are well described by the RDT channel, yet show systematic family-level differences in gradient shape, motivates the use of $(\beta, \kappa, \mathrm{AUC})$ as a richer diagnostic of compression geometry than any single scalar summary.

\begin{figure*}[t]
    \centering
    \includegraphics[width=\textwidth]{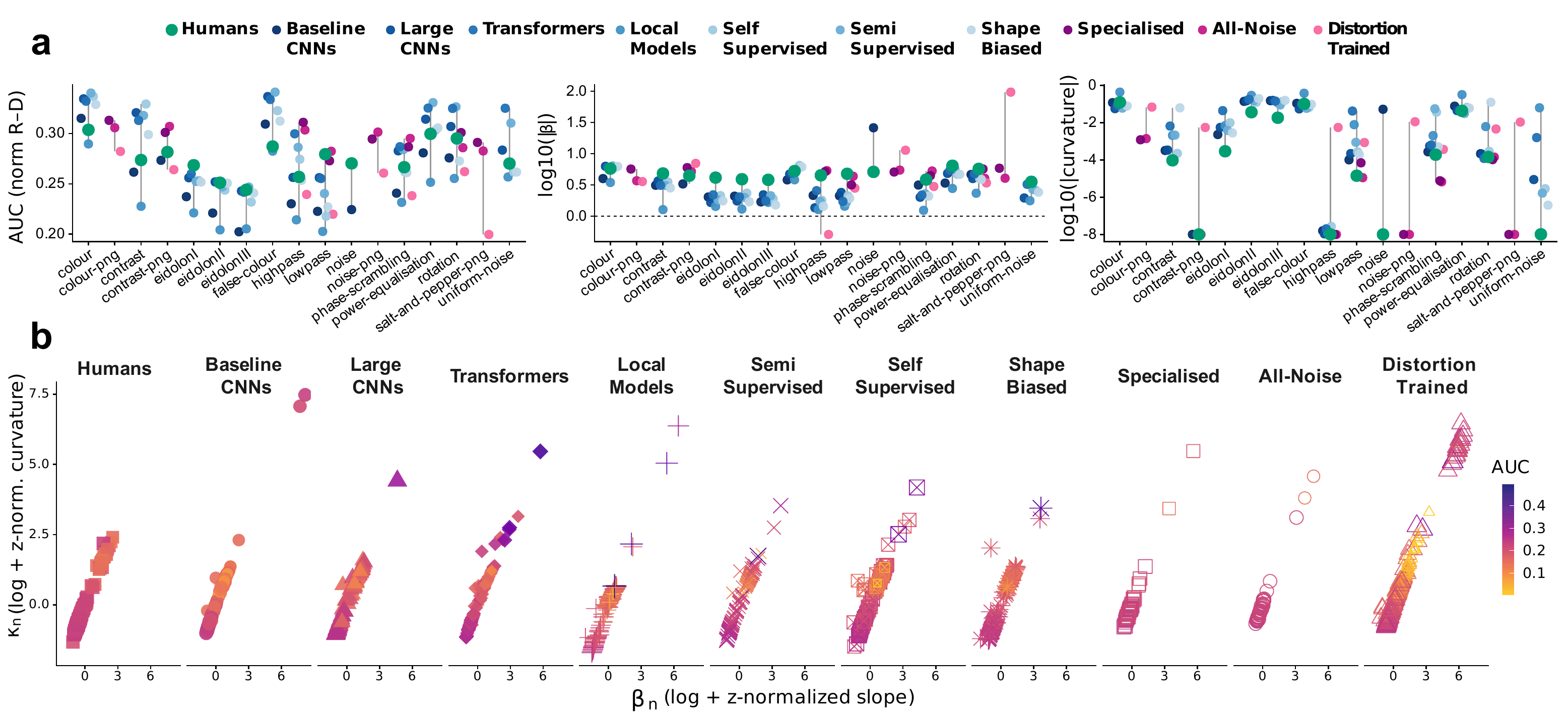}
    \caption{
\textbf{Rate--distortion signatures across perturbation experiments and system families.}
(a) For each experiment/condition (x-axis; abbreviations as shown), points show systems’ RD summaries: normalized rate--distortion efficiency (AUC; left), $\log_{10}|\beta|$ (middle), and $\log_{10}|\kappa|$ (right), grouped by family/regime (legend). Values are computed per experiment$\times$condition block and visualized to highlight between-family structure across contexts.
(b) Global view of the behavioral trade-off landscape: block-level $\beta_n$ (log- and z-normalized slope) vs.\ $\kappa_n$ (log- and z-normalized curvature), faceted by family/training-regime. Marker color encodes AUC. Across families, points form coherent bands with similar orientation but different intercepts, consistent with a shared accuracy--geometry coupling and family/regime-dependent \emph{location shifts} in RD space.}
    \label{fig:heatmap}
\end{figure*}

\subsection*{Architecture families occupy systematically different regions of RD space}

Across 73 matched perturbation contexts (unique experiment$\times$condition blocks for which both human and model data were available) non-human model families consistently occupy shifted regions of RD space relative to humans, with reliably steeper slopes and higher curvature that persist after controlling for accuracy differences between systems (Figure~\ref{fig:heatmap}a; Supplementary Table~\ref{tab:rd_accuracy_stats}). LocalModels showed the largest curvature offset ($\Delta\kappa = 1.46$, approximately $28.8\times$ human curvature, $p_{\mathrm{FDR}} < 10^{-11}$), and ShapeBiased, SelfSupervised, SemiSupervised, and LargeCNNs all exhibited significantly higher $\kappa$ than humans (median $\Delta\kappa \in [0.115, 0.340]$, folds $\approx 1.30\times$ to $2.19\times$; all $q < 0.02$). Vision Transformers were the sole family not distinguishable from humans in curvature ($\Delta\kappa = -0.014$, $q = 0.18$), suggesting that attention-based architectures may implement a qualitatively different compression geometry from convolutional networks. Across all trained families, the separation from an untrained ResNet-50 control (10 random seeds; $\kappa \approx 1.4 \times 10^6$, approximately $19{,}000\times$ human curvature) confirms that this geometric variation is bounded within a learned compression regime qualitatively distinct from random weights (Supplementary Note~7).

Critically, these geometric dissociations are not explained by accuracy differences. CLIP, for example, shows no significant accuracy difference from humans ($\Delta\mathrm{Acc} = 0.025$, $q = 0.11$) yet has reliably larger slope ($\Delta\beta = 0.140$, $q < 10^{-3}$) and curvature ($\Delta\kappa = 0.280$, $q < 10^{-4}$). Block fixed-effects regressions confirm this dissociation formally. After controlling for within-block accuracy, architecture-family indicators remain reliably positive predictors of both $\beta$ and $\kappa$ (Table~\ref{tab:rd_accuracy_stats}), demonstrating that RD geometry captures systematic, family-structured variation in compression strategy that is invisible to performance-based evaluation. Interaction tests further show no evidence that families differ in the form of the accuracy--geometry relationship, indicating that architectures primarily differ by their location in RD space rather than by changes in how performance maps onto geometry.

\subsection*{Training interventions shift compression geometry along dissociable axes}

Robustness-oriented training interventions move models along different axes of the RD trade-off surface, revealing a fundamental dissociation. Interventions that shift compression geometry toward the human regime reduce accuracy and efficiency, while interventions that improve accuracy and efficiency simultaneously push curvature further from the human regime - a trade-off that is entirely invisible to performance-based evaluation (Table~\ref{tab:regime_summary}; Figure~\ref{fig:heatmap}; Table~\ref{tab:geirhos_regime_vs_humans}).

\begin{table}[!htbp]
\centering
\small
\setlength{\tabcolsep}{6pt}
\renewcommand{\arraystretch}{1.3}
\caption{\textbf{Training regime effects on RD geometry relative to humans.} Directional summary of how each training regime shifts slope ($\beta$), curvature ($\kappa$), and accuracy relative to human observers. Arrows indicate the direction of the difference (model minus humans): $\uparrow$ higher, $\downarrow$ lower, $\rightarrow$ closer to human level, $\approx$ not significantly different. See Table~\ref{tab:geirhos_regime_vs_humans} for full
statistics.}
\label{tab:regime_summary}
\begin{tabular}{lcccp{4.5cm}}
\toprule
\textbf{Regime} & \textbf{Accuracy} & \textbf{Slope $\beta$} &
\textbf{Curvature $\kappa$} & \textbf{Interpretation} \\
\midrule
Baseline CNNs &
$\approx$ human &
$\uparrow$ steeper &
$\uparrow$ higher &
High information cost and abrupt compression transitions, qualitatively unlike biological vision. \\
Distortion-trained &
$\downarrow$ lower &
$\rightarrow$ closer &
$\rightarrow$ closer &
More human-like trade-off geometry, but at a cost to task accuracy. \\
All-noise / Specialised &
$\uparrow$ higher &
$\approx$ human &
$\downarrow$ below human &
Higher efficiency and accuracy but curvature moves past the human regime; compression transitions become over-smoothed. \\
\bottomrule
\end{tabular}
\end{table}

Classically pretrained baseline CNNs (GoogLeNet, ResNet-152, VGG-19) remain geometrically separated from humans across all matched blocks, showing reliably larger $\beta$ and $\kappa$ and reduced RD efficiency ($\Delta\mathrm{AUC} \approx -0.02$, $q \leq 10^{-5}$), even in conditions where accuracy differences between models and humans are small. Standard ImageNet training therefore produces a qualitatively different compression geometry from biological vision, independent of task performance.

Distortion-trained models partially close this geometric gap, showing smaller yet significant offsets in both $\beta$ and $\kappa$ relative to humans ($\Delta\beta = 0.188$, $q = 1.31 \times 10^{-5}$), consistent with a shift toward the human region of RD space. However, this geometric movement comes at a cost. Distortion training reduces both accuracy and AUC relative to humans ($q < 10^{-2}$ and $q < 10^{-4}$ respectively), indicating that training on degraded distributions produces more human-like compression geometry, but less efficient overall performance.

Multi-corruption (all-noise) and single-corruption (specialised) regimes show the opposite pattern. Both outperform humans in accuracy ($q < 10^{-5}$) and modestly exceed humans in AUC, yet their curvature shifts past humans in the opposite direction ($\Delta\kappa < 0$, $q < 10^{-3}$ for both), while slope differences become indistinguishable from humans after correction. Robustness training can therefore improve accuracy and efficiency while simultaneously moving the curvature of the compression frontier further from the human regime. This training side-effect is entirely invisible to accuracy-based evaluation, but recoverable from RD geometry.

Together, these regime-specific shifts produce a structured global RD landscape. Across all systems and perturbation contexts, families form coherent bands in the $(\beta_n, \kappa_n)$ plane with similar orientation but displaced intercepts (Figure~\ref{fig:heatmap}b), consistent with a shared accuracy--geometry coupling whose intercept - not slope - is determined by training regime. Slope and curvature are not interchangeable within this landscape, as regimes can approach humans along one axis while diverging along the other, and improvements in AUC efficiency are not consistently associated with proximity to the human compression regime in $(\beta_n, \kappa_n)$ space. This axis-specificity motivates treating $(\beta, \kappa, \mathrm{AUC})$ as complementary diagnostics rather than redundant summaries of a single underlying dimension.

\subsection*{Behavioral RD signatures are grounded in internal representational geometry}

Prior work has shown that when rate-distortion trade-offs are architecturally explicit (i.e., variational autoencoders), the compression geometry directly shapes latent representational structure in predictable ways~\citep{d2025geometry}. We investigate whether analogous structure can be recovered from behavioral confusion data alone in systems where compression is implicit rather than architecturally specified. To test this, we conducted four complementary analyses on 15 ModelZoo networks evaluated across four distortion experiments at graded severities (contrast, uniform noise, EidolonI, and phase scrambling; 31 conditions total): (i) correlation of the MAP-inferred cost matrix $\rho$ with activation RDMs across five processing depths, (ii) partial Spearman correlation controlling for accuracy to isolate signature--geometry relationships independent of task performance, (iii) cross-model analysis testing whether architecture-level differences in signatures predict differences in representational geometry, and (iv) trajectory analysis testing whether signatures predict the dynamics of representational reorganization as distortion severity increases.

\begin{figure*}[t]
\centering
\includegraphics[width=\textwidth]{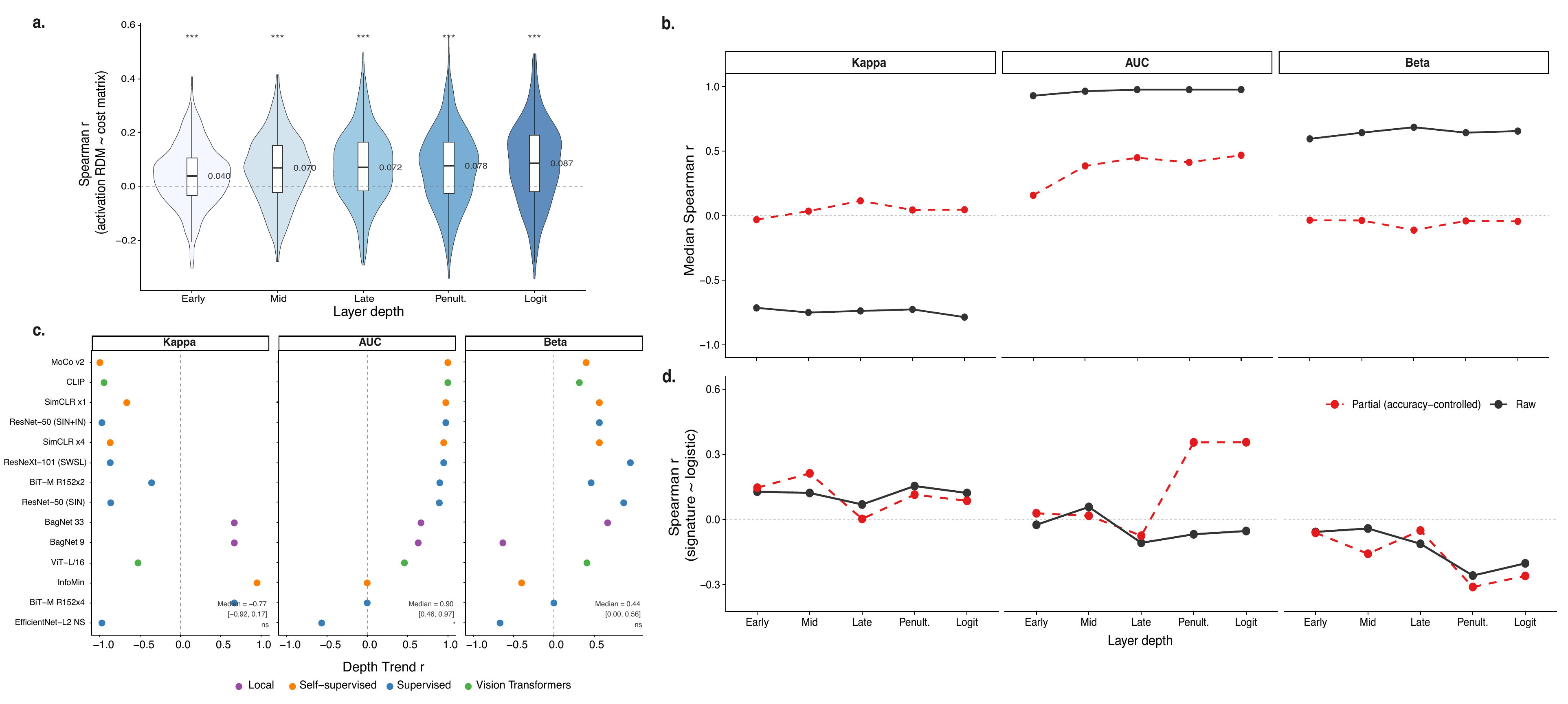}
\caption{\textbf{Behavioral RD signatures are grounded in internal
representational geometry.}
\textbf{(a)} Spearman correlation between the activation RDM and the MAP-inferred cost matrix $\rho$ at each layer depth (violin plots; each point is one model$\times$experiment$\times$condition). \textbf{(b)} Raw (solid) and accuracy-controlled partial (dashed) correlation between mean representational dissimilarity and each RD signature, by layer depth. Panels share the same y-axis. \textbf{(c)} Per-model depth trend in signature--dissimilarity alignment (Spearman correlation between layer index and alignment strength), colored by architecture family. \textbf{(d)} Raw (solid) and partial (dashed) correlation between each RD signature and a logistic nonlinearity index quantifying the abruptness of representational reorganization under degradation, by layer depth. Across all panels, AUC --- but not $\kappa$ or $\beta$ --- tracks internal representational geometry independently of accuracy, with this relationship specific to deep, categorically organized layers.}
\label{fig:rsa}
\end{figure*}

\paragraph{The behaviorally inferred cost structure aligns with internal representations and strengthens with processing depth.} 
For each model, condition, and layer depth, we extracted class-mean activation centroids, computed a $16 \times 16$ RDM using cosine dissimilarity, and correlated its upper triangle with the MAP-inferred cost matrix $\rho$ using Spearman rank correlation. Correlations were positive across all 15 models at the penultimate layer (median $r = +0.082$; all individually significant after controlling false discovery rate using the
Benjamini--Hochberg procedure (BH; ~\citep{benjamini1995controlling}), indicating that class pairs the system treats as costly to confuse are also more dissimilar in internal representational space. Critically, this alignment was not uniform across the processing hierarchy: it increased monotonically from $r = +0.040$ at early layers to $r = +0.087$ at the logit layer, with all five depths significant after BH--FDR correction ($p_{\mathrm{FDR}} < 0.001$), confirming that behavioral cost structure is most precisely encoded in the deeper, more categorically organized layers of the network (Figure~\ref{fig:rsa}a; Supplementary Note~5, Supplementary Tables~\ref{tab:supp_layer_wilcoxon}--\ref{tab:supp_depth_trend}. A per-model breakdown confirming the alignment is not driven by any single architecture is provided in Supplementary Figure~\ref{fig:supp_rho_per_model}.

\paragraph{AUC tracks representational geometry independently of accuracy.}
We next asked which RD signatures predict how strongly the activation RDM covaries with distortion severity-- that is, whether signatures track the steady-state geometry of representations across perturbation
conditions. Raw Spearman correlations between mean RDM dissimilarity and each signature across severity levels were strong for all three signatures (median $|r| \geq 0.60$). However, this reflects a shared confound: as distortion severity increases, accuracy, mean pairwise representational dissimilarity, and all three signatures change together. Applying partial Spearman correlation, to remove the linear contribution of accuracy, revealed a sharp dissociation. The AUC correlation survived accuracy control (partial $r = +0.413$ at penultimate; $p_{\mathrm{FDR}} < 0.001$), whereas $\kappa$ and $\beta$ collapsed to near zero (partial $r = +0.044$ and $-0.041$ respectively; both $p_{\mathrm{FDR}} > 0.60$). A cross-model analysis confirmed that this signal generalizes across architectures. At mid through logit layers, models with higher AUC also showed higher mean representational dissimilarity at matched distortion levels (median cross-model Spearman $r = +0.43$ at mid,
$+0.46$ at logit), whereas $\kappa$ showed the opposite direction at all depths (median $r = -0.16$ to $-0.36$). Therefore, AUC captures information about representational geometry that is genuinely independent of task performance, while $\kappa$ and $\beta$ reflect the shared trajectory of behavioral and representational change across severity rather than an independent link to internal structure (Figure~\ref{fig:rsa}b; raw and partial correlations for all signatures and layers are provided in Supplementary Table~\ref{tab:supp_layer_wilcoxon}; cross-model correlations testing generalizability across architectures are reported in Supplementary Table~\ref{tab:supp_cross_model}).

\paragraph{AUC--geometry alignment is specific to deep categorical representations.}
The layer profile of the accuracy-controlled AUC correlation directly reveals where in the processing hierarchy compression efficiency is encoded. At early and intermediate layers, partial $r$ between mean representational dissimilarity and AUC was negligible ($r \leq 0.017$ at early through late layers). It emerged specifically at the penultimate and logit layers (partial $r = +0.413$ and $+0.356$ respectively; $p_{\mathrm{FDR}} < 0.05$ at both depths). A formal depth trend test confirmed this gradient was systematic across models (one-sample Wilcoxon on per-model trend $r$: median $= +0.897$, 95\% CI $[+0.462,\ +0.975]$, $p_{\mathrm{FDR}} = 0.012$; 11 of 14 models with positive trend), whereas $\kappa$ showed no such gradient (median trend $r = -0.766$, $p_{\mathrm{FDR}} = 0.107$; depth-invariant). AUC therefore indexes a property specific to the categorical, decision-level representations encoded in deep layers (Figure~\ref{fig:rsa}c; Supplementary Note~5, Supplementary Tables~\ref{tab:supp_layer_wilcoxon}--\ref{tab:supp_depth_trend}).

\paragraph{AUC predicts the abruptness of representational
reorganization, specifically at deep layers.}
We further asked whether RD signatures predict not just the level but the dynamics of representational change under degradation. For each system, layer, and experiment, we computed a representational change trajectory $\delta_k = 1 - \mathrm{Spearman}(\mathrm{RDM}_k, \mathrm{RDM}_{\mathrm{baseline}})$, where the baseline RDM was computed from undistorted colour images, providing a universal, experiment-independent reference. A logistic curve fitted to each trajectory yielded a nonlinearity index $k_{\mathrm{fit}}$ (higher values: more threshold-like collapse; lower values: gradual reorganization), which measures the abruptness of the nonlinear trajectory. In the raw analysis, no signature significantly predicted the nonlinearity index ($p_{\mathrm{FDR}} > 0.05$ for all), as accuracy masked the effect. After partial correlation removing accuracy, AUC emerged as a significant predictor specifically at the two deepest layers (partial $r = +0.355$ at penultimate, $+0.356$ at logit; $p_{\mathrm{FDR}} < 0.05$ at both), while remaining near zero at early through late layers (partial $r \leq 0.028$). Neither $\kappa$ nor $\beta$ predicted representational nonlinearity after accuracy control at any depth (all $p_{\mathrm{FDR}} > 0.07$). The same deep-layer representations that most faithfully encode the behaviorally inferred cost structure are therefore also those whose reorganization dynamics are best predicted by AUC (Figure~\ref{fig:rsa}d; Supplementary Note~6, Supplementary Tables~\ref{tab:supp_delta_rdm_layers}--\ref{tab:supp_delta_rdm_experiments}).

\paragraph{$\kappa$ indexes behavioral compression abruptness, not representational geometry.}
The pattern across all four analyses converges on a single interpretive conclusion: whereas AUC captures the quality and organisation of learned categorical representations, $\kappa$ captures how abruptly the system's behavioral compression transitions as distortion increases - a property of the distortion type's statistical structure rather than of internal representational geometry. This interpretation is corroborated by the structure of $\rho$ itself, since high-$\kappa$ systems have more concentrated, lower-entropy cost profiles and lower SVD effective rank, linking the behavioral signature to the geometry of the inferred cost matrix rather than to representational organisation (Supplementary Note~4, Supplementary Table~\ref{tab:supp_rho_structure}; see also Section~\ref{sec:kappa_degradation}).

\subsection*{AUC predicts accuracy degradation trajectory shape, while $\kappa$ shows a distortion-type dependent effect}
\label{sec:kappa_degradation}

To test whether RD signatures predict the shape of accuracy trajectories under increasing perturbation severity, we computed a degradation nonlinearity index for each system--experiment pair ($n = 285$; 41 systems $\times$ 11 experiments with sufficient monotone severity gradients; see Methods). Higher values indicate more abrupt, threshold-like accuracy collapse rather than smooth linear degradation.

Systems with higher AUC degraded more smoothly across perturbation severity (Spearman $r = -0.413$). This relationship survived within-experiment demeaned analysis removing all between-experiment confounds ($t = -5.66$, $p = 3.7 \times 10^{-8}$), confirming that AUC independently predicts degradation trajectory shape above and beyond accuracy ($\Delta$pseudo-$R^2 = 0.327$ in noise-corruption experiments; LRT $p < 0.0001$; Supplementary Note~8, Supplementary Table~\ref{tab:supp_degradation}, Figure~\ref{fig:supp_degradation}a--c). $\kappa$ showed a positive pooled association with degradation nonlinearity (Spearman $r = +0.381$), but this did not survive within-experiment demeaning ($t = 1.85$, $p = 0.066$), indicating partial dependence on between-experiment variation in $\kappa$ range. An experiment-type moderation analysis revealed a significant sign reversal across perturbation classes (LRT $\chi^2(1) = 36.17$, $p < 0.0001$), where for within noise-corruption experiments, higher $\kappa$ predicted more abrupt accuracy degradation ($\beta_{\mathrm{within}} = +0.039$, $p < 0.0001$),
whereas within signal-manipulation experiments the relationship reversed ($\beta_{\mathrm{within}} = -0.054$, $p = 0.0002$). This pattern replicated across both corpora and was consistent across individual experiments (9/11 in the predicted direction; binomial $p = 0.033$; full moderation statistics, corpus-split results, and sensitivity analyses in Supplementary Note~8, Supplementary Table~\ref{tab:supp_degradation}). $\beta$ did not independently predict degradation nonlinearity after accuracy control (all $p > 0.14$; $\Delta$pseudo-$R^2 = 0.001$; Supplementary Table~\ref{tab:supp_degradation}).

\section{Discussion}
\label{sec:discussion}
The efficient coding framework has long provided a principled account of why biological sensory systems are organized as they are, predicting receptive field statistics from natural image statistics, accounting for the tuning properties of early visual neurons, and explaining the universal law of generalization as a consequence of optimal compression under resource constraints~\citep{attneave1954informational,barlow1961possible,simoncelli2001natural,sims2018efficient}. What has remained elusive is a framework that moves beyond asking whether a system is consistent with efficient coding to characterizing the specific \textit{geometry} of its compression strategy --- how steeply and how abruptly it trades fidelity for efficiency, and whether these geometric properties are grounded in the internal representational structure that implements them. The present work provides such a model-agnostic framework by treating stimulus-response behavior as an effective communication channel and extracting geometric signatures of the rate--distortion frontier. We validate this framework using deep neural networks as a ground truth test bed with fully accessible internal representations. Our results establish that although biological and artificial visual systems implement the same lossy compression principle, they occupy systematically different regions of the RD frontier. Architecture families form stable, accuracy-independent clusters in $\beta$-$\kappa$ space, and training regimes reveal that this geometry is not fixed but movable along dissociable axes, with interventions that improve accuracy and efficiency doing so at the cost of human-like curvature, and interventions that recover human-like curvature doing so at the cost of efficiency. Critically, these behavioral signatures are not merely descriptive, as the same geometric properties that separate humans from artificial systems track the geometry of internal representations across the visual processing hierarchy. This positions rate--distortion geometry as a bridge between behavioral psychophysics and the neural implementation of perceptual compression, with direct implications for understanding how biological vision maintains robustness under degradation and for identifying the computational principles that distinguish it from artificial systems.

\paragraph{What the human-DNN dissociation reveals about biological compression.}
The training regime results establish a striking constraint on the relationship between compression efficiency and compression geometry that no current artificial training objective satisfies simultaneously. Distortion training moves models toward the human regime in $(\beta, \kappa)$ space but at the cost of reduced AUC. Meanwhile, robustness-oriented training (all-noise and specialised regimes) improves accuracy and AUC while pushing $\kappa$ further from the human regime. Human observers, by contrast, occupy a region of RD space in which AUC and $\kappa$ are jointly configured in a way that none of the 18 models or four training regimes tested achieves: smooth, low-curvature compression transitions alongside high overall efficiency. That this configuration is not simply a consequence of training on natural statistics is evidenced by the untrained ResNet control, which occupies a degenerate region of RD space entirely outside the range of any trained family, and by the failure of models trained on natural image distributions to approach the human cluster in $(\beta_n, \kappa_n)$ space. The dissociation is therefore not a matter of scale or data quality but of compression strategy. Biological visual systems have discovered --- perhaps through evolution and development --- a configuration of the rate--distortion trade-off that simultaneously minimizes abruptness and maximizes representational efficiency, whereas artificial training objectives optimize these properties in opposition. This has concrete implications for computational models of visual cortex. Models that match human behavioral accuracy but diverge in RD geometry implement a different compression strategy and are therefore likely to diverge from biological neural representations in predictable ways, specifically at the level of categorical geometry in deep visual areas. The axis-specificity of human-likeness provides a principled criterion for identifying which architectural or training innovations bring artificial systems closer to the biological compression regime along which dimension.

\paragraph{A two-level dissociation between compression efficiency and compression abruptness.}
The representational geometry analyses reveal that the three behavioral signatures capture qualitatively different aspects of compression strategy, and that this difference is mechanistically grounded. Geometrically, slope $\beta$ captures the typical marginal cost of reducing distortion along the traced frontier, while curvature $\kappa$ captures how non-uniform that marginal cost is across the frontier, describing the difference between a frontier that either rises steadily or abruptly transitions from flat to a sharp rise. AUC --- the integrated area under the rate--distortion frontier --- independently tracks the quality and organization of internal categorical representations. Its correlation with penultimate-layer representational dissimilarity survives accuracy control, deepens monotonically with processing depth, and generalizes across 11 of 14 architectures tested. This identifies AUC as a behavioral signature of representational quality rather than a more sensitive version of accuracy. This parallels the finding that in generative models with explicit rate-distortion objectives, the compression trade-off geometry directly shapes latent representational structure~\citep{d2025geometry}, but extends it to
discriminative systems where the compression geometry must be inferred from behavior rather than read off from the architecture. Specifically, two systems can match on accuracy while differing in AUC, and that difference predicts how their internal categorical geometry is organized. AUC further predicts the \textit{dynamics} of representational reorganization under degradation. After accuracy control, systems with higher AUC show more abrupt representational reorganization at deep layers (penultimate and logit), while degrading more smoothly at the behavioral level. This apparent inversion reflects that AUC indexes representational quality. Systems
with well-organised categorical geometry undergo sharper representational transitions when that geometry is disrupted, while their behavioral accuracy degrades gracefully precisely because those representations are robustly organised.
$\kappa$, by contrast, does not independently predict representational geometry after accuracy control, and its raw correlation with representational dissimilarity is equally strong at early feature layers and at the penultimate layer. This depth-invariance is the key as it indicates that $\kappa$ tracks a property present throughout the entire visual hierarchy --- the shared trajectory along which behavioral and representational change co-evolve as distortion severity increases --- rather than a property specific to learned categorical structure. What $\kappa$ captures is the abruptness of the compression transition imposed by the distortion type itself, namely how steeply the effective channel must reorganize as compression demands increase. This is a property that reflects the statistical structure of the distortion rather than the system's learned categorical geometry. This interpretation is consistent with $\kappa$'s role as a predictor of behavioral degradation abruptness (the shape of accuracy-versus-severity trajectories), its depth-invariant raw correlation with representational change, and the observation that 
the internal geometry of $\rho$ (specifically its row entropy and SVD effective rank) tracks $\kappa$ strongly across systems, with concentrated, low-entropy cost profiles in high-$\kappa$ systems. This pattern is further consistent with the experiment-type moderation in the behavioral degradation analysis, where $\kappa$ predicts abrupt accuracy collapse specifically in noise-corruption experiments that impose monotonically increasing compression demands, but not in signal-manipulation experiments where severity changes stimulus structure qualitatively rather than intensifying compression pressure (see Section~\ref{sec:kappa_degradation}). Independent evidence for this organizational structure in $\rho$ comes from analysis of its directional asymmetry, examining whether a system confuses class $A$ for $B$ more readily than $B$ for $A$. This shows that artificial networks concentrate directional errors into a small number of dominant collapse directions, whereas humans distribute them more broadly, and that this concentration itself predicts RD efficiency and curvature above and beyond accuracy~\citep{caglar2026directional}.
$\beta$ occupies a more limited role in this dissociation. It does not independently predict representational geometry or its degradation dynamics at any layer after accuracy control, suggesting that the marginal cost of compression, unlike its abruptness or its overall efficiency, carries comparatively little signal about internal representational structure once shared accuracy variance is removed. $\beta$ behaves as a less stable, position-dependent version of the AUC signal, by tracking the local slope at the current operating point rather than the full frontier geometry, making it sensitive to where on the severity gradient the system is operating rather than to the overall quality of its categorical representations.
Together, AUC and $\kappa$ constitute complementary signatures that characterize compression strategy at two distinct levels. AUC captures how efficiently the internal representations that implement compression are organized, while $\kappa$ captures how abruptly the behavioral consequences of compression change as input quality degrades. This is precisely the mechanistic counterpart of the axis-specificity observed behaviorally. Training interventions that move $\kappa$ toward the human regime (distortion training) leave AUC depressed, while interventions that move AUC toward or beyond the human regime (robustness training) leave $\kappa$  elevated, because the two signatures are grounded in different representational properties -- one in the categorical organization of deep representations, the other in the depth-invariant statistical structure of the distortion itself -- and a single training objective that reshapes one does not automatically reshape the other. The behavioral dissociation between $\beta$-$\kappa$ and AUC observed across training regimes and the representational dissociation between $\kappa$ and AUC observed in the internal representational linking analyses are therefore two views of the same underlying fact, namely, compression efficiency and compression abruptness are governed by different aspects of a system's computation, and only biological vision appears to have arrived at a configuration that jointly optimizes both.

\paragraph{Biological relevance and a specific prediction for neural data.}
The present demonstration of the relationship between RD signatures and internal representations was conducted entirely in artificial systems, where such internal representations are fully accessible. This is a deliberate design choice that establishes the framework's validity before applying it to biological neural recordings, where representations can only be partially observed. Nevertheless, several features of the results speak directly to the neural implementation of perceptual compression in biological vision. The layer-depth gradient for AUC --- negligible at early and intermediate layers, significant only at penultimate and logit representations --- is structurally similar to the primate ventral stream hierarchy, in which categorical abstraction increases progressively from V1 through V4 to inferotemporal cortex ~\citep{dicarlo2012does,yamins2014performance}. This parallel generates a specific and falsifiable prediction for neural data, suggesting that AUC computed from behavioral confusion matrices should correlate with representational dissimilarity structure in inferotemporal cortex and late ventral stream areas more strongly than in early visual cortex. Meanwhile, $\kappa$, which shows depth-invariance across all five tested processing stages, should show no such cortical gradient. This consistency holds across architectures spanning CNNs, transformers, and self-supervised models, and extends to shape-biased ResNets trained on Stylized ImageNet \citep{geirhos2019imagenet} indicating that AUC reflects a general property of learned categorical representations rather than an artifact of any particular architectural family. These predictions can be tested directly using representational similarity analysis of fMRI responses or multi-unit electrophysiology \citep{khaligh2014deep,yamins2014performance}, provided that behavioral confusion matrices are collected from the same subjects and stimuli. 
For distortion type, the present results identify noise-type perturbations like uniform noise and phase scrambling as the best candidates, since AUC's prediction of representational reorganization dynamics was strongest for perturbations that impose monotonically increasing compression demands rather than qualitative stimulus changes. It is important to note that for the RD geometry approach, the within-subject design is essential as it enables the behaviorally-inferred RD signatures to be computed from the same individuals whose neural geometry is measured, providing the most direct test of whether the compression strategy inferred from behavior is instantiated in the underlying neural representations. 

\paragraph{Limitations and future directions.}
Three limitations circumscribe the present conclusions. First, the rate quantities reported here are explicitly behavioral. $I(X;Y)$ is computed from output confusion statistics and reflects an effective stimulus-response rate under a specified task and cost geometry, not mutual information in internal representations. The internal representation linking analyses provide partial validation that behavioral RD geometry recovers internal representational structure, but they do not constitute a direct measurement of internal coding efficiency. The most direct remedy is precisely the within-subject neural validation outlined above, in which behavioral RD signatures and neural representational geometry are measured in the same individuals. Second, the present study uses a controlled perturbation suite and a 16-way categorization task, and how stable $(\beta, \kappa, \mathrm{AUC})$ remain under changes in dataset semantics, label granularity, or long-tailed class distributions remains an open question. Third, the specific predictions for cortical area specificity and distortion-type dependence proposed above have not yet been tested empirically and require dedicated neuroimaging or electrophysiological data.
A natural extension of the present framework is to generative models and autoencoders, where the rate-distortion trade-off is architecturally explicit and the compression geometry can be directly manipulated via the $\beta$ hyperparameter. D'Amato et al.~\citep{d2025geometry} showed that this explicit trade-off distorts the latent geometry of generative models in predictable ways. The present behavioral approach offers a complementary route to characterizing compression geometry from the outside - from confusion patterns alone - without requiring access to latent representations. Testing whether the behavioral RD signatures recovered from such models' output confusions correspond to the latent geometry shapes that are directly measurable would provide a further validation of the framework's ability to recover internal structure from behavior. Another extensions of this framework include application to adversarially trained and certified-robust models, where adversarial training can be understood as altering the effective distortion geometry \citep{madry2017towards} by emphasizing worst-case confusions, and RD signatures would quantify whether robustness under adversarial perturbations reflects smoother trade-offs (reduced $\kappa$) or sharper regime boundaries (increased $\kappa$). Finally, the framework extends naturally to the characterization of atypical compression geometry in clinical populations. The prediction that AUC tracks representational quality degradation and $\kappa$ tracks abruptness of compression transitions under perceptual load generates specific hypotheses about how disrupted efficient coding in conditions such as schizophrenia or frontotemporal dementia \citep{adams2013computational} would manifest in behavioral confusion structure, providing a behaviorally accessible measure that could complement neural recording in populations and help reduce cost and patient burden.

\begin{ack}
This work was supported in part through the Minerva computational and data resources and staff expertise provided by Scientific Computing and Data at the Icahn School of Medicine at Mount Sinai and supported by the Clinical and Translational Science Awards (CTSA) grant UL1TR004419 from the National Center for Advancing Translational Sciences.
\end{ack}

\section*{Methods}

\subsection*{Participants and neural network models}

\textbf{Human observers.}
Human psychophysical data were obtained from two publicly available repositories. The GEN repository~\citep{geirhos2019imagenet} provides approximately 83,000 forced-choice categorization trials collected under controlled perturbation conditions. The ModelZoo repository~\citep{geirhos2021partial} provides an independent human dataset of approximately 85,000 trials, designed for matched model--human comparisons with consistent stimulus presentation. In both datasets, participants performed a 16-way forced-choice categorization task on natural images under 12 families of controlled perturbations at graded severities. All human data were collected under the experimental protocols described in the original benchmark publications~\citep{geirhos2019imagenet,geirhos2021partial}. No new human data were collected for the present study.

\textbf{Neural network models.}
Eighteen pretrained deep neural network models were evaluated, spanning four broad categories. \textit{Supervised CNNs} include standard ImageNet-trained baselines (GoogLeNet~\citep{szegedy2015going}, ResNet-152~\citep{he2016deep}, VGG-19~\citep{simonyan2015verydeep}), large-scale pretrained models (BiT-M ResNetV2-152x2 and 152x4~\citep{kolesnikov2020big}), and architecturally specialized local-feature networks (BagNet-9, BagNet-33~\citep{brendel2019approximating}). \textit{Transformer and vision--language models} include ViT-L/16~\citep{Dosovitskiy2021}, SWAG-ViT-L/16~\citep{singh2022swag}, and CLIP~\citep{radford2021clip}. \textit{Self- and semi-supervised models} include SimCLR ResNet-50x1 and x4~\citep{chen2020simple}, MoCo v2~\citep{chen2020mocov2}, InfoMin~\citep{tian2020infomin}, SWSL ResNeXt-101~\citep{mahajan2018exploring}, and Noisy Student EfficientNet-L2~\citep{xie2020noisystudent}. \textit{Shape-biased models} include ResNet-50 trained on Stylized ImageNet (SIN) and a mixture of Stylized and standard ImageNet (SIN+IN)~\citep{geirhos2019imagenet}.

In addition to these architecture families, four training regimes were evaluated using ResNet-50 variants trained from scratch on a 16-class ImageNet subset: a \textit{Distortion-trained} regime using a distorted training distribution, a \textit{Specialised} regime using single-distortion training, and an \textit{All-noise} regime using multi-corruption training. For comparison and as a \textit{Baseline} regime, we use the standard supervised ImageNet-trained CCN models GoogLeNet, ResNet-152, and VGG-19 evaluated without any robustness-specific intervention. These regimes enable controlled comparison of how training distribution affects compression geometry independently of architecture.
All ModelZoo models were evaluated using the \texttt{modelvshuman} toolbox~\citep{geirhos2021partial}. GEN repository models were evaluated using their original pretrained checkpoints. For all models, confusion matrices reflect hard (argmax) decisions.

\textbf{Untrained baseline control.}
To anchor the RD space with a principled lower bound on learned compression, we additionally evaluated ten randomly initialized ResNet-50 models (distinct random seeds: 42, 137, 256, 391, 512, 628, 741, 853, 967, 1024) with no training. Weights were drawn from PyTorch default initialization (Kaiming uniform) and no gradient updates were applied. Each seed was evaluated identically to the trained ModelZoo models across all 12 perturbation experiments and their full severity ranges, yielding 850 confusion matrices in total (10 seeds × 85 experiment × condition combinations). The ten seeds quantify variability from random initialization independently of training, allowing RD signatures to be reported as mean ± SD across seeds. Untrained models achieve 6.2\% ± 0.2\% accuracy, consistent with chance performance on the 16-way task and are excluded from all primary human--model comparisons, since they serve only as a reference anchor in the architecture family analysis (see Supplementary Note~7).

\subsection*{Stimulus benchmark}

All systems were evaluated on a 16-way ImageNet-derived categorization task using the generalization benchmark of Geirhos et al.~\citep{geirhos2019imagenet}, which comprises 12 perturbation families: colour/grayscale, contrast, high-pass filtering, low-pass filtering (blur), phase scrambling, power equalisation, opponent colour, rotation, Eidolon I, Eidolon II, Eidolon III, and uniform noise. Each perturbation type is parameterized by distortion strength, yielding a systematic range of out-of-distribution evaluation conditions at graded severities. Full details of stimulus generation and experimental protocols are provided in the original benchmark publication~\citep{geirhos2019imagenet}.

\subsection*{Behavioral channel construction}

For each system $s$ (human observer or neural network model), experiment $e$, and distortion level $d$, we summarize stimulus--response behavior with a $K \times K$ confusion matrix $N^{(s,e,d)}$ of counts, where $N_{ij}$ is the number of trials or images with true class $i$ for which the system responded with class $j$. Row-normalizing yields an empirical conditional distribution
\begin{equation}
C^{(s,e,d)}_{ij} \;=\; \Pr_s(y = j \mid x = i;\, e, d)
\;\approx\;
\frac{N^{(s,e,d)}_{ij}}{\sum_{j'} N^{(s,e,d)}_{ij'}},
\label{eq:channel_from_confusions}
\end{equation}
which we treat as an effective behavioral channel $p_s(y \mid x;\, e, d)$ for all downstream analyses. For human data, $N^{(s,e,d)}$ is obtained by aggregating forced-choice trial responses. For models, we compute one predicted label per image and aggregate hard decisions (argmax predictions) into counts. Confusion matrices therefore reflect categorical decisions rather than probability-averaged responses. All rate--distortion fits and signatures operate on the row-normalized channel $C^{(s,e,d)}$, where off-diagonal structure is analyzed separately, $C_0^{(s,e,d)}$
denotes $C^{(s,e,d)}$ with diagonal entries set to zero

\subsection*{Rate--distortion curve estimation}

\textbf{Inference of distortion geometry.}
Rather than assuming a fixed 0--1 classification loss, we treat the distortion (error-cost) structure as a latent quantity inferred from empirical confusions, following the behavioral rate--distortion approach of Sims~\citep{sims2018efficient}. For each experiment, we infer a cost matrix $\rho \in \mathbb{R}_{\geq 0}^{K \times K}$ by maximum a posteriori (MAP) optimization under a structured prior that regularizes symmetric, asymmetric, and diagonal components of $\rho$ separately. Given a candidate $\rho$, the corresponding optimal channel $q_\lambda(y \mid x)$ is obtained by Blahut--Arimoto fixed-point updates~\citep{blahut1972computation,arimoto1972algorithm}, and the fit of $\rho$ is evaluated by comparing this implied channel to the empirical behavior. The resulting inferred cost matrix $\rho$ captures graded structure in the pattern of confusions beyond a binary correct/incorrect loss. Posterior uncertainty around the MAP estimate can be approximated locally via a Laplace approximation based on the Hessian at the optimum.

\textbf{Rate--distortion frontier construction.}
Given the inferred $\rho$, we trace a discrete approximation to the rate--distortion frontier by scaling $\rho$ with an inverse-temperature parameter $\lambda > 0$ and computing the corresponding optimal channel:
\begin{equation}
q_\lambda(y \mid x) \;\propto\; p(y) \exp\{-\lambda\,\rho(x, y)\},
\label{eq:optimal_channel_lambda}
\end{equation}
iterated jointly with $p(y) = \sum_x p(x)\,q_\lambda(y \mid x)$
until convergence.
For a log-spaced grid of $\lambda$ values ($\lambda \in [10^{-1}, 10^3]$), we compute the mutual information $R_k = I(X; Y)$ and expected distortion $D_k = \mathbb{E}[\rho(X, Y)]$ under an empirical class prior, yielding rate--distortion points $\{(D_k, R_k)\}_k$ for each system and experiment. These points constitute a curve-level summary of the trade-off geometry implied by the behavioral channel $C^{(s,e,d)}$ and the inferred distortion structure $\rho$.

\subsection*{Rate--distortion signatures}

From each estimated rate--distortion frontier $\{(D_k, R_k)\}_{k=1}^{K}$, we extract three compact summary statistics.

\textbf{Slope ($\beta$).} Local slopes are estimated by finite differences, $s_k = (R_{k+1} - R_k)/(D_{k+1} - D_k)$, after sorting by $D$ and removing duplicate $D$ values. Slope is summarized as $\beta \equiv \mathrm{median}_k\, s_k$. The mean is reported as a secondary summary where noted.

\textbf{Curvature ($\kappa$).} Curvature is quantified as the dispersion of local slopes, $\kappa \equiv \mathrm{Var}_k(s_k)$, which is zero for an exactly linear trade-off and increases as the marginal information cost of reducing distortion varies more strongly across the frontier.

\textbf{Area under the RD curve (AUC).} Efficiency is computed as the area under the $R(D)$ curve over the traced distortion range using trapezoidal integration after filtering non-finite points.

Where used, within-experiment normalized signatures ($\beta_n$, $\kappa_n$) are obtained by applying a log transform followed by z-scoring within each experiment, to place slope and curvature on comparable scales across perturbation families.

\subsection*{Analyses linking internal representations to RD signatures}

\textbf{Activation centroid extraction and RDM construction.}
For each model$\times$layer$\times$condition, we computed class-mean activation centroids by averaging activations across all images of each of the 16 categories. Activations were extracted by registering forward hooks at five depth points per architecture: early (first residual block or equivalent), mid (second block), late (third block), penultimate (global average pool, normalization layer, or equivalent pre-classifier representation), and logit (pre-softmax features). Hook targets were architecture specific. For ResNet-family models (including BagNets, SimCLR, SIN-trained variants, and ResNeXt), hooks were registered at \texttt{layer1}, \texttt{layer2}, \texttt{layer3}, \texttt{avgpool}, and the input to \texttt{fc}; or BiT (ResNetV2), at \texttt{stages[0--2]}, \texttt{norm}, and \texttt{head}; for EfficientNet, at compound blocks 1, 3, and 5, \texttt{global\_pool}, and \texttt{classifier} input; for ViT-L/16 and SWAG ViT-L/16, at transformer encoder layers 0, 7, and 15, the layer normalization \texttt{encoder.ln}, and the classification head; for CLIP and MoCo/ InfoMin (contrastive architectures without an accessible class logit), at four depth points only (no logit hook). Centroids were computed as the mean activation vector across all images of each class for a given condition, yielding a $16 \times d$ matrix per (model, layer, condition). A $16 \times 16$ representational dissimilarity matrix (RDM) was constructed from each centroid matrix using cosine dissimilarity ($1 - \cos\theta$). This approach follows the representational similarity analysis (RSA)
framework~\citep{kriegeskorte2008representational}, which characterizes
representational geometry by comparing pairwise dissimilarity structures
across conditions, layers, or systems.

\textbf{Scope of conditions.}
The layer-resolved analysis covered four distortion experiments with graded severity levels---contrast (8 levels), uniform noise (8 levels), EidolonI (8 levels), and phase scrambling (7 levels)---yielding 31 conditions per model, selected to provide sufficient severity gradients for trajectory analyses. Penultimate-layer centroids for all 15 models across all four experiments were extracted on GPU (NVIDIA A100) using the \texttt{modelvshuman} toolbox. The additional four non-penultimate layers required a dedicated extraction pass using registered PyTorch forward hooks.

\textbf{RDM--$\rho$ correlation.}
For each model$\times$layer$\times$experiment$\times$condition, the upper triangle of the activation RDM was correlated with the corresponding upper triangle of the MAP-inferred $\rho$ matrix using Spearman rank correlation. Statistical significance was assessed using BH--FDR correction across all model$\times$condition pairs within each layer. The layer-depth gradient was tested by computing, for each model, the Spearman correlation between layer index (0--4) and median Spearman $r$ across conditions (depth trend $r$), and then applying a one-sample Wilcoxon signed-rank test against zero across the 14 models with at least four depth points.

\textbf{Partial correlation controlling for accuracy.}
To isolate representational geometry effects independent of shared severity-driven variance, we applied partial Spearman correlation. For each model $\times$ layer $\times$ experiment, we computed the Spearman correlation between mean pairwise representational dissimilarity (the mean of the upper-triangular entries of the activation RDM) and each RD signature ($\kappa$, AUC, $\beta$) across severity conditions, after removing the linear contribution of accuracy ($A$, proportion correct) from both variables. Specifically, we rank-transformed all variables, regressed accuracy ranks from the signature ranks and from mean pairwise representational dissimilarity ranks, and computed Pearson correlation on the residuals. This yields partial Spearman $r$ values that quantify the signature--geometry association independent of the shared accuracy--severity trajectory. BH--FDR correction was applied across all model$\times$experiment pairs within each layer$\times$signature combination.

\textbf{Cross-model analysis.}
As a complementary test of generalizability across architectures rather than within a single model's severity gradient, we computed Spearman correlations between mean pairwise representational dissimilarity and each RD signature \emph{across the 15 models} at each layer$\times$experiment$\times$condition. This tests whether models with higher AUC (or $\kappa$) also have systematically higher or lower mean representational dissimilarity at the same distortion level. Only conditions with at least 8 models contributing finite values were included.

\textbf{$\delta$(RDM) trajectory and nonlinearity analysis.}
To characterize the dynamics of representational change across severity, we computed a representational change trajectory for each model$\times$layer$\times$experiment. A universal baseline RDM was constructed from class-mean centroids extracted from undistorted colour images (condition \texttt{cr} from the colour experiment), providing an experiment-independent reference reflecting the model's representational geometry on natural, unperturbed input. The representational change at severity level $k$ was then defined as:
\begin{equation}
\delta_k \;=\; 1 - \mathrm{Spearman}(\mathrm{RDM}_k,\,
\mathrm{RDM}_{\mathrm{baseline}}).
\label{eq:delta_rdm}
\end{equation}
$\delta_k = 0$ indicates no change from the baseline geometry;
$\delta_k = 1$ indicates complete decorrelation. A logistic curve
$\delta(s) = L / (1 + e^{-k_{\mathrm{fit}}(s - s_0)})$ was fitted to
each trajectory as a function of log-severity using nonlinear least
squares (scipy \texttt{curve\_fit}; bounds: $L \in [0, 2]$,
$k_{\mathrm{fit}} \in [0, 20]$). The logistic slope $k_{\mathrm{fit}}$ serves as the nonlinearity index where higher values indicate more abrupt, threshold-like representational reorganization and lower values indicate gradual and smooth change. For the contrast experiment, severity was defined as the inverse of the contrast level (severity $= 101 - c$, where $c \in \{1, 3, 5, 10, 15, 30, 50, 100\}$), so that higher values correspond to greater degradation, consistent with the other experiments. Fit convergence was confirmed for all 288 model$\times$layer$\times$experiment pairs. As a model-free backup, the variance of second differences of $\delta$ across severity levels was also computed. This measure requires no curve-fitting assumptions and yielded consistent directional results. The relationship between each RD signature and $k_{\mathrm{fit}}$ was tested using raw and partial Spearman correlation (controlling for mean accuracy), with BH--FDR correction within each signature across all layers.

\subsection*{Degradation nonlinearity analysis}

To test whether RD signatures predict the shape of accuracy-versus-severity trajectories, we derived a nonlinearity index for each system and perturbation experiment from accuracy measurements across graded severity levels. The nonlinearity index quantifies deviation from a linear degradation trajectory, where a value of zero indicates perfectly linear accuracy decline with severity, and larger values indicate more abrupt, threshold-like collapse. Only perturbation experiments with at least four distinct severity levels and a monotone mean accuracy gradient were included. Excluded experiments (rotation, colour, false-colour, and power-equalisation) lacked monotone severity gradients suitable for trajectory shape analysis. Spearman rank correlations between each RD signature and the nonlinearity index were computed across all included system--experiment pairs, controlling for mean accuracy level. BH--FDR correction was applied across all signature--experiment combinations. An experiment-type moderation analysis distinguished noise-corruption perturbations (uniform noise, blur, Eidolon families) from signal-manipulation perturbations (contrast, phase scrambling) to assess whether the $\kappa$--degradation coupling varied by perturbation type.

\subsection*{Statistics}

Rate--distortion summary metrics ($\beta$, $\kappa$, AUC, and accuracy) are computed for each system within each experimental context, defined as a matched experiment$\times$condition block. Within each block, different systems evaluated on the same stimuli constitute repeated measurements, enabling paired block-aware inference via within-block contrasts. False discovery rates are controlled using the BH--FDR procedure throughout. Effect sizes are reported as median blockwise differences and rank-biserial correlations.

\textbf{ModelZoo comparisons.} In the ModelZoo dataset, all humans and models have data for matched stimuli across experiments, so all family-level comparisons are performed on an identical block set ($n_\text{blocks} = 73$).

\textbf{GEN training-regime comparisons.} In the GEN repository, training-regime datasets are not fully crossed. Each regime comparison is therefore restricted to the subset of matched blocks for the two systems being contrasted, and $\Delta$ is defined as (regime $-$ humans) within shared blocks. Because the distortion-trained regime contains multiple independently trained instances per block (up to 20 models), instance-level values are collapsed to a single per-block median before computing regime-level contrasts.

\textbf{Model-by-model comparisons.} For each model, within-block differences from humans ($\Delta m_b = m_{b,\text{model}} - m_{b,\text{human}}$) are tested for deviation from zero using a two-sided paired Wilcoxon signed-rank test across blocks.

\textbf{Family-level comparisons.} Higher-level separations across model families or regimes are tested by aggregating to a single median value per block$\times$family before performing paired Wilcoxon signed-rank tests for planned contrasts.

\textbf{Accuracy-adjusted regressions.} To test whether RD geometry differs across families beyond accuracy, we fit within-block fixed-effects regressions using block-demeaned variables: $\beta_r \sim \mathrm{acc}_r + \mathrm{family}_r + \mathrm{block}$, where $\mathrm{acc}_r$ and $\beta_r$ (and analogously $\kappa_r$) are residualized by subtracting within-block means, with humans as the reference family. Whether the coupling between accuracy and RD geometry differs by family is tested via nested-model interaction tests comparing the additive and interaction models using ANOVA for nested OLS models. These tests distinguish shifts in location in RD space (family offsets) from changes in the accuracy--geometry relationship (interaction effects).

\bibliographystyle{plainnat}
\bibliography{RateDistortion}


\appendix


\renewcommand{\thesection}{A\arabic{section}}
\setcounter{section}{0}

\FloatBarrier
\section*{Supplementary Information}
\renewcommand{\figurename}{Supplementary Figure}
\renewcommand{\tablename}{Supplementary Table}
\setcounter{figure}{0}
\setcounter{table}{0}

\subsection*{Supplementary Note 1: RD slope dynamics across 
perturbation severity}

Certain perturbations — including phase scrambling, severe noise, and extreme contrast reduction — systematically increase compression demands for all systems. Across experiments, increasing perturbation strength steepens the inferred RD slope signature $\beta$ (paired tests across experiments, $p \approx 0.02$), indicating that higher information rates are required to maintain accuracy as inputs become less reliable. While this monotonic difficulty--$\beta$ trend has been documented in human generalization~\citep{sims2018efficient}, our results show that many artificial vision models exhibit the same direction of change. However, matching the direction of the effect does not imply matching its \emph{dynamics}, since different systems can reach high-compression regimes either gradually or abruptly as conditions worsen.

To make these dynamics explicit, we analyze a perturbation setting where severity is explicitly parametrized by a continuous corruption scale, focusing on a uniform-noise experiment in which increasing noise level acts as a within-experiment stress test on the effective channel. For each system and noise level, we compute a slope $\beta$ by regressing the log-probability of off-diagonal confusions against the inferred cost of error $\rho$, pooling all off-diagonal class pairs. Because some off-diagonal entries can be zero at particular noise levels, we apply additive pseudocount (Laplace) smoothing only for this visualization step before row-normalizing:
\begin{equation}
\tilde{C}_{ij}
\;=\;
\frac{N_{ij} + \alpha}{\sum_{j'} (N_{ij'} + \alpha)},
\label{eq:laplace_smooth}
\end{equation}
where $N_{ij}$ denotes the raw confusion count for true class $i$ and response $j$, and $\alpha = 0.5$. This smoothing is applied only to stabilize log-probabilities in the severity visualization and is not used in RD fitting or in the primary RD signatures reported in the main text.

As shown in Supplementary Figure~\ref{fig:BetaSlopes}, human observers exhibit a gradual steepening of $\beta$ with increasing noise. As visual input becomes less reliable, the effective channel increases the marginal information cost of accuracy in a smooth, distributed manner. Baseline CNNs show a markedly sharper increase in $|\beta|$ across the same corruption range, consistent with a more abrupt transition into a high-compression regime. Specialised and all-noise networks can exhibit especially steep slopes at intermediate-to-high noise levels, indicating that even when models match the direction of the severity effect, they can diverge substantially from humans in how sensitively and how abruptly their generalization geometry changes across the corruption scale.

\begin{figure}[!htbp]
    \centering
    \includegraphics[width=0.8\linewidth]{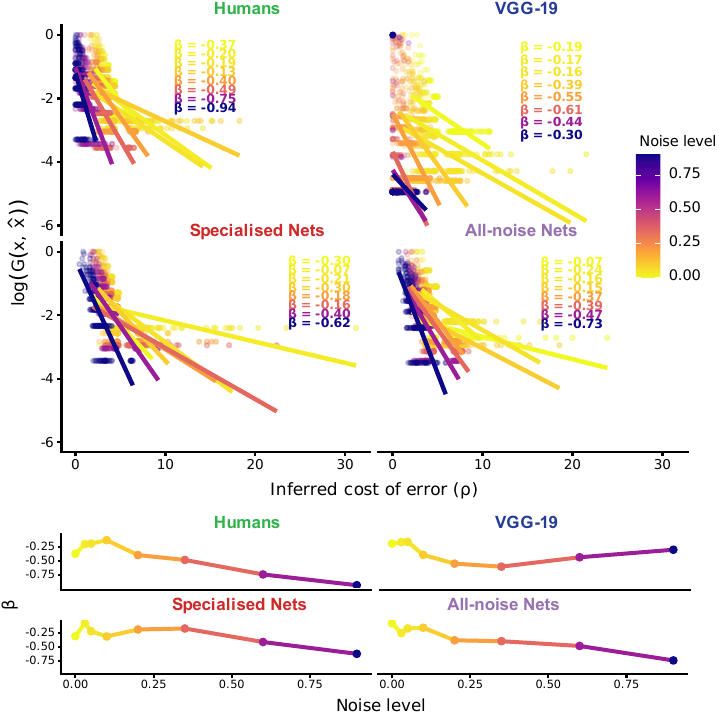}
    \caption{
        \textbf{Effects of noise on generalization slopes across humans and models.}
        Top panels show, for each system, the relationship between inferred error cost ($\rho$) and the log-probability of off-diagonal confusions, with colored lines denoting different uniform-noise levels. Each line’s slope defines $\beta$ for that noise level. Confusion counts are stabilized with an additive pseudocount ($\alpha=0.5$) before row-normalization for this visualization. Bottom panels plot $\beta$ as a function of noise level, revealing gradual severity dependence in humans and sharper, more noise-sensitive transitions in baseline CNNs, with specialised and all-noise networks exhibiting intermediate patterns.
    }
    \label{fig:BetaSlopes}
\end{figure}
\FloatBarrier

\subsection*{Supplementary Note 2: Goodness-of-fit summary}

Supplementary Table~1 reports RMSE-based goodness-of-fit statistics for each model family across three complementary diagnostics: (i) deviation from the rate-matched exponential generalization curve $G = \exp(-s\rho)$; (ii) observed versus predicted off-diagonal confusion probabilities; and (iii) deviation between the empirical binned generalization gradient and its best-fit exponential form $ae^{-\beta d}$. Lower RMSE indicates better agreement with the corresponding rate--distortion prediction. These values are summarized in the main text and full per-family values are provided in Supplementary Table~1.

\begin{table}[!htbp]
\caption{\textbf{RMSE-based goodness-of-fit by model family.}
For each model family we report the number of fitted files and
three complementary RMSE measures: (i) deviation from the
rate-matched exponential generalization curve
$G = \exp(-\beta\,\rho)$, (ii) observed vs.\ predicted
off-diagonal confusion probabilities, and (iii) deviation between
the empirical binned generalization gradient and its best-fit
exponential form $a\,e^{-\beta d}$. Lower RMSE indicates better
agreement with the corresponding rate--distortion prediction.}
\label{tab:rmse_family}
\centering
\small
\setlength{\tabcolsep}{5pt}
\renewcommand{\arraystretch}{1.2}
\begin{tabular}{lrrrr}
\toprule
Model Family & \#files &
\makecell{Generalization\\curve fit\\(median RMSE)} &
\makecell{Confusion\\probability fit\\(RMSE)} &
\makecell{Empirical\\gradient fit\\(mean RMSE)} \\
\midrule
All-noise          & 53   & 0.007 & 0.010 & 0.004 \\
Baseline           & 243  & 0.049 & 0.014 & 0.031 \\
Distortion-trained & 1220 & 0.052 & 0.011 & 0.029 \\
Humans             & 154  & 0.023 & 0.018 & 0.008 \\
LargeCNNs          & 146  & 0.030 & 0.027 & 0.029 \\
LocalModels        & 146  & 0.491 & 0.032 & 0.043 \\
SelfSupervised     & 292  & 0.048 & 0.025 & 0.038 \\
SemiSupervised     & 146  & 0.034 & 0.024 & 0.041 \\
ShapeBiased        & 146  & 0.090 & 0.031 & 0.025 \\
Specialised        & 53   & 0.009 & 0.011 & 0.006 \\
VisionTransformers & 219  & 0.025 & 0.023 & 0.047 \\
\bottomrule
\end{tabular}
\end{table}

\FloatBarrier

\subsection*{Supplementary Note 3: Detailed statistical tables}

Supplementary Tables~2 and 3 provide full model-by-model and regime-level paired comparison statistics referenced in the main text. Supplementary Table~2 reports within-block paired comparisons (model $-$ humans) across $n_\text{blocks} = 73$ matched ModelZoo contexts, together with fixed-effects regression results testing whether RD geometry adds information beyond accuracy. Supplementary Table~3 reports Geirhos training-regime comparisons across matched perturbation blocks, including block counts, median within-block differences, rank-biserial effect sizes, and BH--FDR adjusted $q$-values for each metric. In both tables, $\Delta$ denotes the blockwise median difference (model $-$ humans), fold-changes are reported as $10^\Delta$ for log-scale parameters, and all $p$-values are two-sided.

\FloatBarrier
\begin{table*}[!htbp]
\centering
\scriptsize
\setlength{\tabcolsep}{3.5pt}
\renewcommand{\arraystretch}{1.15}
\caption{\textbf{Model-level paired comparisons vs humans and fixed-effects tests of RD geometry beyond accuracy.}
\textbf{Panel A} reports paired, within-block differences (model $-$ humans) across $n_{\mathrm{blocks}}=73$ matched contexts using two-sided Wilcoxon signed-rank tests; $q$ denotes BH--FDR-adjusted $p$-values (adjusted within each metric across the 15 model contrasts).
\textbf{Panel B} reports fixed-effects regressions of RD parameters on within-block accuracy and architecture-family indicators with block fixed effects (block coefficients omitted).}
\label{tab:rd_accuracy_stats}

\smallskip
\noindent\textbf{Panel A: Model-by-model paired comparisons vs humans (Wilcoxon signed-rank; $n_{\mathrm{blocks}}=73$).}\\[-0.3em]
\begin{adjustbox}{width=\textwidth}
\begin{tabular}{lrrrrrrrr}
\toprule
Model & $\Delta$Acc & $q_{\mathrm{Acc}}$ & $\Delta\beta$ (fold) & $q_{\beta}$ & $\Delta\kappa$ (fold) & $q_{\kappa}$ & $\Delta\mathrm{AUC}$ & $q_{\mathrm{AUC}}$ \\
\midrule
BagNet-9 & -0.339 & $<10^{-4}$ & +0.432 (×2.71) & $<10^{-4}$ & +1.900 (×79.43) & $<10^{-4}$ & -0.0554 & $<10^{-4}$ \\
BagNet-33 & -0.104 & $<10^{-4}$ & +0.308 (×2.03) & $<10^{-4}$ & +0.911 (×8.15) & $<10^{-4}$ & -0.0283 & $<10^{-4}$ \\
ViT-L/16 & +0.104 & $<10^{-4}$ & +0.062 (×1.15) & $0.13$ & -0.095 (×0.80) & $0.42$ & +0.0355 & $<10^{-4}$ \\
EffNet-L2 (NoisyStudent) & +0.115 & $<10^{-4}$ & +0.068 (×1.17) & $0.13$ & -0.114 (×0.77) & $0.33$ & +0.0458 & $<10^{-4}$ \\
SWAG ViT-L/16 & +0.104 & $<10^{-4}$ & +0.012 (×1.03) & $0.28$ & -0.185 (×0.65) & $0.96$ & +0.0429 & $<10^{-4}$ \\
SimCLR R50x4 & +0.080 & $0.007$ & +0.031 (×1.08) & $0.32$ & -0.088 (×0.82) & $0.53$ & +0.0322 & $0.15$ \\
BiT-M R152x2 & +0.048 & $0.045$ & +0.167 (×1.47) & $<10^{-4}$ & +0.119 (×1.32) & $0.021$ & +0.0191 & $0.041$ \\
CLIP & +0.025 & $0.11$ & +0.140 (×1.38) & $<10^{-4}$ & +0.280 (×1.91) & $<10^{-4}$ & +0.0131 & $0.31$ \\
BiT-M R152x4 & +0.051 & $0.13$ & +0.126 (×1.34) & $0.008$ & +0.083 (×1.21) & $0.015$ & +0.0130 & $0.18$ \\
ResNet50 (SIN) & +0.013 & $0.30$ & +0.141 (×1.38) & $<10^{-4}$ & +0.196 (×1.57) & $<10^{-4}$ & +0.0157 & $0.96$ \\
ResNet50 (SIN+IN) & -0.008 & $0.37$ & +0.225 (×1.68) & $<10^{-4}$ & +0.479 (×3.01) & $<10^{-4}$ & +0.0023 & $0.38$ \\
ResNeXt101 (SWSL) & +0.013 & $0.64$ & +0.100 (×1.26) & $0.002$ & +0.244 (×1.75) & $<10^{-4}$ & +0.0084 & $0.96$ \\
SimCLR R50x1 & +0.011 & $0.64$ & +0.203 (×1.59) & $<10^{-4}$ & +0.370 (×2.34) & $<10^{-4}$ & +0.0229 & $0.58$ \\
MoCoV2 & +0.005 & $0.67$ & +0.112 (×1.29) & $0.005$ & +0.316 (×2.07) & $<10^{-4}$ & +0.0142 & $0.96$ \\
InfoMin & +0.001 & $0.72$ & +0.123 (×1.33) & $0.004$ & +0.275 (×1.88) & $<10^{-4}$ & +0.0076 & $0.83$ \\
\bottomrule
\end{tabular}
\end{adjustbox}

\smallskip
\noindent\textbf{Panel B: Fixed-effects regressions testing whether RD parameters add information beyond accuracy.}\\[-0.3em]
\begin{adjustbox}{width=0.92\textwidth}
\begin{tabular}{lrrrr}
\toprule
Predictor & $\hat{\beta}$ (SE) & $p_{\beta}$ & $\hat{\kappa}$ (SE) & $p_{\kappa}$ \\
\midrule
Accuracy (within-block) & -0.516 (0.080) & $<10^{-4}$ & -3.286 (0.174) & $<10^{-4}$ \\
LargeCNNs & +0.186 (0.050) & $2.01\times 10^{-4}$ & +0.590 (0.109) & $<10^{-4}$ \\
VisionTransformers & +0.144 (0.047) & $0.002$ & +0.711 (0.103) & $<10^{-4}$ \\
SemiSupervised & +0.163 (0.050) & $0.001$ & +0.748 (0.109) & $<10^{-4}$ \\
SelfSupervised & +0.106 (0.046) & $0.021$ & +0.615 (0.099) & $<10^{-4}$ \\
ShapeBiased & +0.196 (0.050) & $<10^{-4}$ & +0.619 (0.109) & $<10^{-4}$ \\
LocalModels & +0.228 (0.053) & $<10^{-4}$ & +0.765 (0.116) & $<10^{-4}$ \\
\bottomrule
\end{tabular}
\end{adjustbox}

\smallskip
\noindent{\footnotesize\textbf{Notes.} A \emph{block} is a matched experimental context (experiment $\times$ perturbation condition). All paired tests compare models to humans \emph{within the same block}. In Panel A, $\Delta$ denotes the within-block median difference (model $-$ humans) and ``fold'' is $10^{\Delta}$ for log-parameters. In Panel B, humans are the reference family. Block fixed-effects absorb all block-specific shifts (coefficients omitted for brevity). Reported $p$-values are two-sided.}
\end{table*}
\FloatBarrier

\begin{table*}[t]
\centering
\scriptsize
\setlength{\tabcolsep}{3.5pt}
\renewcommand{\arraystretch}{1.15}
\caption{\textbf{GEN repository: model-/regime-level paired comparisons vs humans across matched perturbation blocks.}
For each model/regime, we compute within-block differences relative to humans (model $-$ humans) across matched contexts (\emph{blocks}; experiment $\times$ condition), then test whether the median within-block difference differs from zero using two-sided Wilcoxon signed-rank tests.
We report the block count ($n_{\mathrm{blocks}}$), median within-block difference ($\Delta$), rank-biserial effect size ($r_{\mathrm{rb}}$), and BH--FDR-adjusted $q$ within each metric (across the 6 model contrasts).
For log-parameters, $\Delta\beta$ and $\Delta\kappa$ are in $\log_{10}$ units (fold-change $=10^{\Delta}$).}
\label{tab:geirhos_regime_vs_humans}
\begin{adjustbox}{width=\textwidth}
\begin{threeparttable}
\begin{tabular}{l r r r r r r r r r r r r r}
\toprule
& & \multicolumn{3}{c}{Accuracy} & \multicolumn{3}{c}{$\beta$ (log$_{10}$)} & \multicolumn{3}{c}{$\kappa$ (log$_{10}$)} & \multicolumn{3}{c}{AUC} \\
\cmidrule(lr){3-5}\cmidrule(lr){6-8}\cmidrule(lr){9-11}\cmidrule(lr){12-14}
Model / regime & $n_{\mathrm{blocks}}$
& $\Delta$Acc & $q_{\mathrm{Acc}}$ & $r_{\mathrm{rb}}$
& $\Delta\beta$ & $q_{\beta}$ & $r_{\mathrm{rb}}$
& $\Delta\kappa$ & $q_{\kappa}$ & $r_{\mathrm{rb}}$
& $\Delta\mathrm{AUC}$ & $q_{\mathrm{AUC}}$ & $r_{\mathrm{rb}}$ \\
\midrule
All-noise         & 33 & +0.177    & $3.52\times 10^{-6}$ & +0.996 & +0.0408 & $8.88\times 10^{-2}$ & +0.358 & -0.145 & $4.23\times 10^{-4}$  & -0.715 & +0.0112  & $7.93\times 10^{-3}$ & +0.544 \\
Specialised       & 33 & +0.120    & $3.52\times 10^{-6}$ & +0.971 & +0.0913 & $1.08\times 10^{-1}$ & +0.323 & -0.177 & $3.20\times 10^{-3}$  & -0.590 & +0.00471 & $4.35\times 10^{-2}$ & +0.405 \\
VGG-19            & 81 & -0.0162   & $3.77\times 10^{-3}$ & -0.398 & +0.292  & $3.64\times 10^{-9}$ & +0.778 & +0.368 & $3.50\times 10^{-11}$ & +0.881 & -0.0209  & $2.83\times 10^{-7}$ & -0.699 \\
Distortion-trained & 35 & -0.0798  & $6.07\times 10^{-3}$ & -0.559 & +0.188  & $1.31\times 10^{-5}$ & +0.863 & +0.253 & $9.98\times 10^{-5}$  & +0.775 & -0.0256  & $2.57\times 10^{-5}$ & -0.835 \\
ResNet-152        & 81 & -0.0104   & $5.15\times 10^{-2}$ & -0.259 & +0.292  & $4.46\times 10^{-10}$& +0.833 & +0.335 & $1.47\times 10^{-9}$  & +0.792 & -0.0197  & $1.72\times 10^{-5}$ & -0.569 \\
GoogLeNet         & 81 & -0.000154 & $8.32\times 10^{-2}$ & -0.222 & +0.299  & $1.80\times 10^{-8}$ & +0.736 & +0.331 & $1.47\times 10^{-9}$  & +0.788 & -0.0222  & $1.02\times 10^{-6}$ & -0.653 \\
\bottomrule
\end{tabular}
\begin{tablenotes}
\footnotesize
\item \textbf{Notes.} A \emph{block} is an (experiment $\times$ condition) context. All comparisons are paired to humans \emph{within the same block}.
Reported $\Delta$ values are blockwise medians of (model $-$ humans). $q$ values are BH--FDR adjusted \emph{within each metric} across the six model contrasts.
\textit{Distortion-trained} comprises 20 independently trained instances. Blockwise distortion-trained summaries are aggregated across instances before block-collapsed testing, and overlap with humans is $n_{\mathrm{blocks}}=35$.
\end{tablenotes}
\end{threeparttable}
\end{adjustbox}
\end{table*}

\subsection*{Supplementary Note 4: Cost matrix structure analysis}
\label{sec:supp_rho_structure}

The MAP-inferred cost matrix $\rho$ encodes the full pattern of confusion costs inferred from behavior. Beyond using $\rho$ to compute RD signatures, we characterized its internal geometry directly using two mathematically appropriate measures for cost matrices, namely SVD effective rank and row entropy.

\textbf{SVD effective rank} (participation ratio from singular values) quantifies how many dimensions carry the cost structure: $\mathrm{PR}_{\sigma} = (\sum \sigma_i)^2 / \sum \sigma_i^2$, where $\{\sigma_i\}$ are the singular values of $\rho$. Values near 1 indicate a single dominant dimension and values near 16 indicate diffuse, high-dimensional cost structure. Eigendecomposition was avoided because $\rho$ is not constrained to be positive semi-definite by the MAP estimation procedure, and universal negative eigenvalues (8--15 of 16 across all 1,249 matrices) render eigenspectrum analysis methodologically inappropriate.

\textbf{Row entropy} quantifies how concentrated each category's cost profile is. For each row of $\rho$, entries are normalized to a probability distribution over target categories and Shannon entropy (nats) is computed. Values are averaged across all 16 rows. Low row entropy indicates that a few category pairs dominate the cost structure. High row entropy indicates diffuse, uniform costs.

\begin{figure}[!htbp]
\centering
\includegraphics[width=0.72\linewidth]{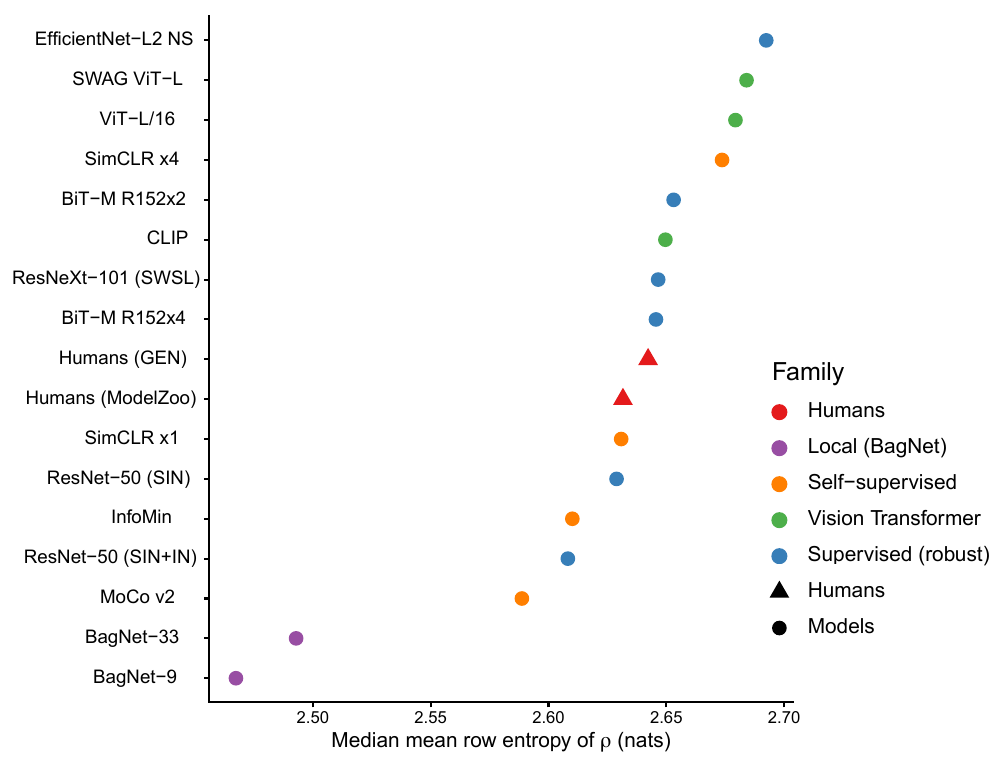}
\caption{\textbf{Per-system cost matrix row entropy.}
Median mean row entropy of the MAP-inferred cost matrix $\rho$ for each model and for human observers (GEN and ModelZoo corpora shown separately as triangles), ordered by entropy value. Lower entropy indicates a more concentrated cost profile in which a small number of category pairs dominate the confusion cost structure. Both human datasets fall within the model distribution rather than at an extreme, consistent with humans occupying an intermediate position in terms of cost matrix geometry. BagNets show the lowest row entropy, consistent with their high $\kappa$ values and concentrated, locally driven confusion structure.}
\label{fig:supp_row_entropy}
\end{figure}

Both metrics were computed for all 1,249 model$\times$experiment$\times$ condition pairs and correlated with $\kappa$ across all model conditions (Supplementary Table~4). $\kappa$ was strongly predicted by both row entropy (Spearman $r = -0.851$, $p < 10^{-300}$, $n = 1{,}095$) and SVD effective rank ($r = +0.809$, $p < 10^{-250}$, $n = 1{,}095$), while Frobenius distance from a uniform cost matrix showed a negligible association in terms of effect size ($r = -0.079$), despite reaching nominal significance ($p = 8.7 \times 10^{-3}$) owing to the large sample size ($n = 1{,}095$). Per-model Spearman correlations confirmed consistency across architectures: for row entropy, all 15 models showed the expected direction (median per-model $r = -0.831$, one-sample Wilcoxon $p_{\mathrm{FDR}} < 0.001$); for SVD effective rank, 14 of 15 models were consistent (median $r = +0.800$, $p_{\mathrm{FDR}} < 0.001$). A direct comparison of humans and models showed that humans have marginally higher row entropy (human median $= 2.638$, model median $= 2.621$; Wilcoxon $W = 74{,}968$, $p = 0.026$) and SVD effective rank (human median $= 4.793$, model median $= 4.781$; $W = 96{,}418$, $p = 0.004$), placing humans within the model distribution rather than at an extreme (Supplementary Figure~\ref{fig:supp_row_entropy}). These findings connect the behavioral signature $\kappa$ directly to the internal geometry of the inferred compression cost structure where systems with higher $\kappa$ have more concentrated, lower-entropy cost profiles organized around fewer dominant dimensions.

\begin{table}[!htbp]
\caption{\textbf{Cost matrix structure metrics: correlations with $\kappa$ and human--model comparison.}
\textbf{Panel A}: Spearman correlations between cost matrix structure metrics and $\kappa$ across all model$\times$experiment$\times$condition pairs ($n = 1{,}095$), and per-model Wilcoxon test of directional consistency.
\textbf{Panel B}: Human vs.\ model comparison on each metric (Wilcoxon rank-sum; all conditions pooled). Human values (both ModelZoo and GEN corpora combined) are compared against all model conditions.}
\label{tab:supp_rho_structure}
\centering
\small
\setlength{\tabcolsep}{5pt}
\renewcommand{\arraystretch}{1.2}

\vspace{0.4em}
\noindent\textbf{Panel A: Correlations with $\kappa$
($n_{\mathrm{cond}} = 1{,}095$ model conditions).}

\vspace{0.3em}
\begin{tabular}{lrrrrrr}
\toprule
Metric & $r$ (all cond.) & $p$ & Median per-model $r$ &
$n_{\mathrm{models}}$ & $p_{\mathrm{FDR}}$ (Wilcoxon) \\
\midrule
Row entropy (nats)        & $-0.851$ & $<10^{-300}$ & $-0.831$ & 15 &
$<0.001$ \\
SVD effective rank        & $+0.809$ & $<10^{-250}$ & $+0.800$ & 15 &
$<0.001$ \\
Frob.\ dist.\ (uniform)  & $-0.079$ & $8.7\times10^{-3}$ & --- & ---
& --- \\
\bottomrule
\end{tabular}

\vspace{0.8em}
\noindent\textbf{Panel B: Human vs.\ model comparison.}

\vspace{0.3em}
\begin{tabular}{lrrrr}
\toprule
Metric & Human median & Model median & $W$ & $p$ \\
\midrule
SVD effective rank  & 4.793 & 4.781 & 96{,}418 & 0.004 \\
Row entropy (nats)  & 2.638 & 2.621 & 74{,}968 & 0.026 \\
\bottomrule
\end{tabular}
\end{table}

\FloatBarrier

\subsection*{Supplementary Note 5: Full statistics for the layer-resolved internal representational linking analysis}
\label{sec:supp_layer_rsa}

\textbf{Raw and partial correlations by layer.}
Supplementary Table~5 reports, for each RD signature$\times$layer depth combination, the median raw and partial Spearman correlation between the mean pairwise representational dissimilarity and the signature across all model$\times$experiment pairs ($n = 44$--$60$ per layer), together with the number of pairs remaining significant after partial correlation and BH--FDR correction. The partial correlation controls for the linear contribution of mean accuracy, removing the shared severity-driven trajectory. Significance was assessed by one-sample Wilcoxon signed-rank test (H0: median $r = 0$) across all pairs within each signature$\times$layer cell, with BH--FDR correction applied within each signature across layers.

\textbf{Depth trend inference.}
For each model, the Spearman correlation between layer index (0--4, shallow to deep) and median signature - mean pairwise representational dissimilarity correlation across experiments was computed as a depth trend $r$. Supplementary Table~6 reports one-sample Wilcoxon tests on the distribution of per-model trend values, together with bootstrap 95\% confidence intervals on the median trend $r$ (5{,}000 resamples). A significant positive trend indicates that the signature--geometry alignment strengthens with processing depth.

\textbf{Cross-model analysis.}
Supplementary Table~\ref{tab:supp_cross_model} reports the median cross-model Spearman correlation between mean pairwise representational dissimilarity and each signature across the 15 models at each layer$\times$experiment$\times$condition combination. This tests whether models with higher signature values have systematically higher or lower mean representational dissimilarity at the same distortion level---a test of generalizability across architectures rather than across severity levels within one model. All 31 experiment$\times$condition pairs met the inclusion threshold of at least 8 models with finite values.

\begin{table}[!htbp]
\caption{\textbf{Raw and accuracy-controlled (partial) correlations
between mean pairwise representational dissimilarity and RD signatures by layer depth.}
For each signature $\times$ layer cell, raw and partial Spearman $r$ values are medians across all model $\times$ experiment pairs ($n$). Partial correlations remove the linear contribution of mean accuracy from both variables. $n_{\mathrm{sig}}$ is the number of pairs remaining significant after BH--FDR correction of the partial correlation $p$-values. One-sample Wilcoxon significance (H0: median $r = 0$) is indicated by stars: $^{***}p_{\mathrm{FDR}} < 0.001$, $^{**}p_{\mathrm{FDR}} < 0.01$, $^{*}p_{\mathrm{FDR}} < 0.05$, ns: not significant.}
\label{tab:supp_layer_wilcoxon}
\centering
\small
\setlength{\tabcolsep}{4pt}
\renewcommand{\arraystretch}{1.2}
\begin{tabular}{llrrrrrr}
\toprule
Signature & Layer & $n$ & Raw $r$ & Sig. & Partial $r$ & Sig. &
$n_{\mathrm{sig}}$ (partial) \\
\midrule
\multirow{5}{*}{AUC}
 & Early       & 56 & $+0.929$ & $^{***}$ & $+0.158$ & $^{***}$ & 14 \\
 & Mid         & 56 & $+0.964$ & $^{***}$ & $+0.385$ & $^{***}$ & 19 \\
 & Late        & 56 & $+0.976$ & $^{***}$ & $+0.449$ & $^{***}$ & 19 \\
 & Penultimate & 60 & $+0.976$ & $^{***}$ & $+0.413$ & $^{***}$ & 19 \\
 & Logit       & 44 & $+0.976$ & $^{***}$ & $+0.468$ & $^{***}$ & 18 \\
\midrule
\multirow{5}{*}{$\beta$}
 & Early       & 56 & $+0.595$ & $^{***}$ & $-0.035$ & ns & 1 \\
 & Mid         & 56 & $+0.643$ & $^{***}$ & $-0.037$ & ns & 2 \\
 & Late        & 56 & $+0.685$ & $^{***}$ & $-0.112$ & ns & 1 \\
 & Penultimate & 60 & $+0.643$ & $^{***}$ & $-0.041$ & ns & 3 \\
 & Logit       & 44 & $+0.655$ & $^{***}$ & $-0.044$ & ns & 1 \\
\midrule
\multirow{5}{*}{$\kappa$}
 & Early       & 56 & $-0.714$ & $^{***}$ & $-0.031$ & ns & 0 \\
 & Mid         & 56 & $-0.750$ & $^{***}$ & $+0.035$ & ns & 0 \\
 & Late        & 56 & $-0.738$ & $^{***}$ & $+0.115$ & ns & 0 \\
 & Penultimate & 60 & $-0.726$ & $^{***}$ & $+0.044$ & ns & 0 \\
 & Logit       & 44 & $-0.786$ & $^{***}$ & $+0.046$ & ns & 0 \\
\bottomrule
\end{tabular}
\end{table}

\begin{table}[!htbp]
\caption{\textbf{Depth trend inference: per-model Spearman correlation between layer index and signature--mean pairwise representational dissimilarity alignment.}
Trend $r$ is the Spearman correlation between layer index (0 = early, 4 = logit) and median (mean pairwise representational dissimilarity $\sim$ signature) correlation across experiments, computed separately for each model. The one-sample Wilcoxon test tests H0: median trend $r = 0$ across models and the bootstrap 95\% CI (5,000 resamples) bounds the median. $n_{\mathrm{pos}}$ = number of models with positive trend; $n_{\mathrm{sig}}$ = number significant after BH--FDR correction.}
\label{tab:supp_depth_trend}
\centering
\small
\setlength{\tabcolsep}{5pt}
\renewcommand{\arraystretch}{1.2}
\begin{tabular}{lrrrrrrr}
\toprule
Signature & $n_{\mathrm{models}}$ & Median trend $r$ &
95\% CI & $n_{\mathrm{pos}}$ & $n_{\mathrm{sig}}$ &
Wilcoxon $p_{\mathrm{FDR}}$ \\
\midrule
AUC      & 14 & $+0.897$ & $[+0.462,\ +0.975]$ & 11 & 6 &
$0.012^{*}$ \\
$\beta$  & 14 & $+0.436$ & $[\ \ 0.000,\ +0.564]$ & 10 & 0 &
$0.107^{\mathrm{ns}}$ \\
$\kappa$ & 14 & $-0.766$ & $[-0.923,\ +0.167]$ &  4 & 3 &
$0.107^{\mathrm{ns}}$ \\
\bottomrule
\end{tabular}
\end{table}

\begin{table}[!htbp]
\caption{\textbf{Cross-model Spearman correlations between mean RDM dissimilarity and RD signatures by layer depth.}
For each signature $\times$ layer combination, cross-model correlations were computed at each experiment $\times$ condition pair across the 15 models, and the median is reported. $n_{\mathrm{cond}} = 31$ per layer (number of experiment $\times$ condition pairs with $\geq 8$ models contributing finite values). $n_{\mathrm{sig}}$ = number of conditions with $p_{\mathrm{FDR}} < 0.05$ after BH--FDR correction within each signature.}
\label{tab:supp_cross_model}
\centering
\small
\setlength{\tabcolsep}{5pt}
\renewcommand{\arraystretch}{1.2}
\begin{tabular}{llrrrr}
\toprule
Signature & Layer & $n_{\mathrm{cond}}$ & Median cross-model $r$ &
$n_{\mathrm{sig}}$ & \% sig \\
\midrule
\multirow{5}{*}{AUC}
 & Early       & 31 & $+0.116$ & 0 & 0.0\% \\
 & Mid         & 31 & $+0.429$ & 2 & 6.5\% \\
 & Late        & 31 & $+0.420$ & 2 & 6.5\% \\
 & Penultimate & 31 & $+0.329$ & 2 & 6.5\% \\
 & Logit       & 31 & $+0.455$ & 1 & 3.2\% \\
\midrule
\multirow{5}{*}{$\beta$}
 & Early       & 31 & $+0.086$ & 0 & 0.0\% \\
 & Mid         & 31 & $+0.160$ & 0 & 0.0\% \\
 & Late        & 31 & $+0.218$ & 0 & 0.0\% \\
 & Penultimate & 31 & $+0.171$ & 0 & 0.0\% \\
 & Logit       & 31 & $+0.209$ & 0 & 0.0\% \\
\midrule
\multirow{5}{*}{$\kappa$}
 & Early       & 31 & $-0.156$ & 0 & 0.0\% \\
 & Mid         & 31 & $-0.284$ & 1 & 3.2\% \\
 & Late        & 31 & $-0.305$ & 1 & 3.2\% \\
 & Penultimate & 31 & $-0.236$ & 0 & 0.0\% \\
 & Logit       & 31 & $-0.364$ & 0 & 0.0\% \\
\bottomrule
\end{tabular}
\end{table}

\FloatBarrier

\begin{figure}[!htbp]
\centering
\includegraphics[width=0.85\linewidth]{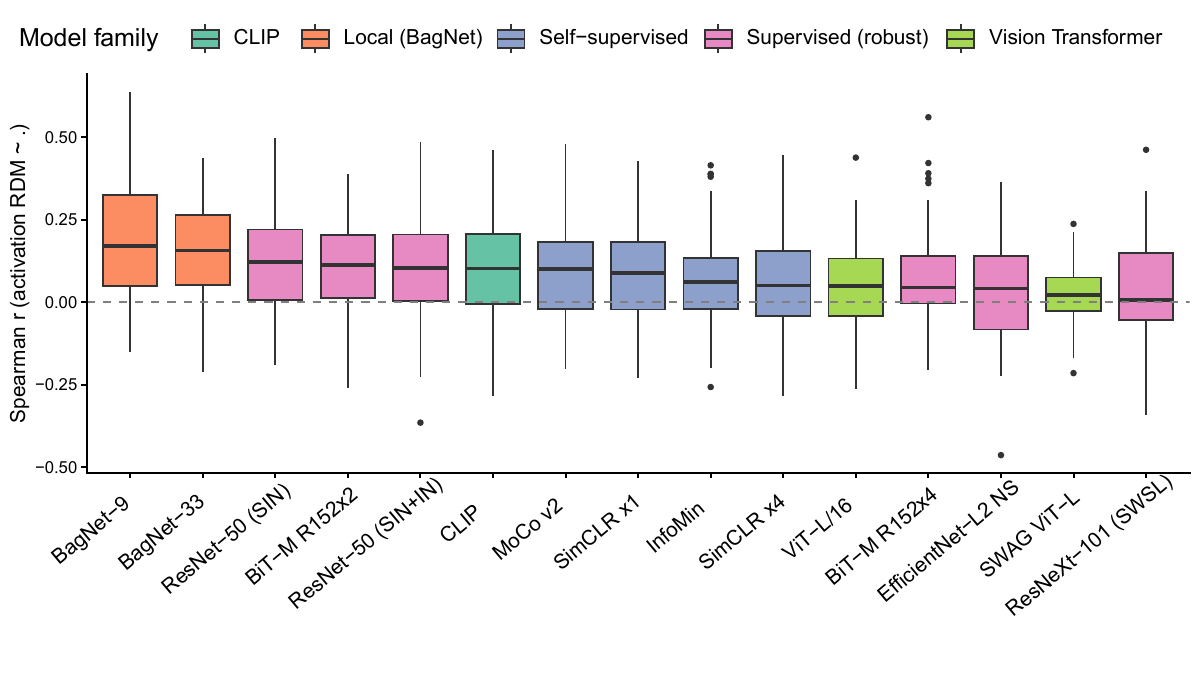}
\caption{\textbf{Per-model alignment between activation RDM and behaviorally inferred cost matrix $\rho$.} Spearman correlation between the penultimate-layer activation RDM and the MAP-inferred cost matrix $\rho$, shown separately for each of the 15 models (box plots; each point is one experiment$\times$condition). Positive correlations indicate that class pairs treated as costly to confuse in behavior are also more dissimilar in internal representational space. All models show positive median correlations, confirming that the
RDM--$\rho$ alignment reported in the main text is not driven by any single model or architecture family.}
\label{fig:supp_rho_per_model}
\end{figure}

\subsection*{Supplementary Note 6: Full statistics and per-experiment breakdown for the $\delta$(RDM) nonlinearity analysis}
\label{sec:supp_delta_rdm}

We provide complete statistics for the representational change trajectory analysis and document the experiment-specific pattern of AUC's prediction of representational nonlinearity.
For each model $\times$ layer $\times$ experiment, a logistic nonlinearity index $k_{\mathrm{fit}}$ was extracted from the $\delta$(RDM) trajectory (see Methods). Raw and accuracy-controlled Spearman correlations between each RD signature and $k_{\mathrm{fit}}$ were computed at each layer depth. BH--FDR correction was applied within each signature across layers. Logistic fits converged for all 288 model $\times$ layer $\times$ experiment pairs at all five layer depths ($100\%$ convergence).

\textbf{Layer-resolved statistics.}
Supplementary Table~8 reports raw and partial Spearman $r$ between each RD signature and the logistic nonlinearity index at each layer depth, together with BH--FDR-corrected significance. No signature predicted representational nonlinearity in the raw analysis at any depth (all $p_{\mathrm{FDR}} > 0.05$), reflecting the masking effect of the shared accuracy signal. After partial correlation removing accuracy, AUC emerged as a significant predictor at the penultimate and logit layers only ($r = +0.355$ and $+0.356$ respectively; $p_{\mathrm{FDR}} < 0.05$ at both depths). This layer specificity directly parallels the layer-wise internal representation results. Neither $\kappa$ nor $\beta$ survived accuracy control at any depth.

\textbf{Per-experiment breakdown at the penultimate layer.}
The AUC--nonlinearity relationship varied systematically across distortion types (Supplementary Table~9). In the raw analysis, AUC showed a positive relationship with representational nonlinearity for noise-type perturbations (uniform noise: $r = +0.500$; phase scrambling: $r = +0.371$) and a negative relationship for stimulus-disrupting perturbations (contrast: $r = -0.036$; EidolonI: $r = -0.154$). This is consistent with the interpretation that $k_{\mathrm{fit}}$ is most meaningful when severity corresponds to graded compression pressure on a fixed distortion type, as in noise addition, rather than qualitative stimulus restructuring, as in eidolon distortions which disrupt global shape while preserving local energy. $\kappa$ showed inconsistent direction across experiments (contrast: $+0.346$; uniform noise: $-0.396$; EidolonI: $+0.093$; phase scrambling: $-0.261$), explaining its near-zero pooled correlation and failure to survive accuracy control.

\textbf{Model-free nonlinearity index.}
As a check that results do not depend on logistic curve fitting, we also computed the variance of second differences of $\delta$ across severity levels as a model-free nonlinearity measure. At the penultimate layer, AUC showed a positive association with this measure after accuracy control ($r = +0.147$, $p = 0.263$), consistent in direction but weaker and non-significant, reflecting the lower sensitivity of the model-free index relative to the fitted logistic slope. $\kappa$ showed $r = +0.164$ (partial, $p = 0.210$) and $\beta$ showed $r = -0.011$ (partial, $p = 0.935$). The directional consistency with the logistic index supports the robustness of the AUC finding.

\begin{table}[!htbp]
\caption{\textbf{$\delta$(RDM) nonlinearity index: raw and partial correlations with RD signatures by layer depth.}
For each signature $\times$ layer combination, Spearman $r$ is computed between the logistic nonlinearity index $k_{\mathrm{fit}}$ and the RD signature mean across severity conditions, across all model $\times$ experiment pairs ($n = 60$ for penultimate; $n = 48$ for logit). Partial $r$ removes the linear contribution of mean accuracy. BH--FDR correction is applied within each signature across layers. Stars: $^{*}p_{\mathrm{FDR}} < 0.05$; ns: not significant.}
\label{tab:supp_delta_rdm_layers}
\centering
\small
\setlength{\tabcolsep}{4pt}
\renewcommand{\arraystretch}{1.2}
\begin{tabular}{llrrrr}
\toprule
Signature & Layer & $n$ & Raw $r$ & Partial $r$ & Sig.\ (partial) \\
\midrule
\multirow{5}{*}{AUC}
 & Early       & 60 & $-0.025$ & $+0.028$ & ns \\
 & Mid         & 60 & $+0.057$ & $+0.017$ & ns \\
 & Late        & 60 & $-0.108$ & $-0.075$ & ns \\
 & Penultimate & 60 & $-0.069$ & $+0.355$ & $^{*}$ \\
 & Logit       & 48 & $-0.054$ & $+0.356$ & $^{*}$ \\
\midrule
\multirow{5}{*}{$\beta$}
 & Early       & 60 & $-0.058$ & $-0.062$ & ns \\
 & Mid         & 60 & $-0.042$ & $-0.158$ & ns \\
 & Late        & 60 & $-0.113$ & $-0.051$ & ns \\
 & Penultimate & 60 & $-0.259$ & $-0.312$ & ns \\
 & Logit       & 48 & $-0.203$ & $-0.261$ & ns \\
\midrule
\multirow{5}{*}{$\kappa$}
 & Early       & 60 & $+0.128$ & $+0.146$ & ns \\
 & Mid         & 60 & $+0.122$ & $+0.212$ & ns \\
 & Late        & 60 & $+0.069$ & $+0.002$ & ns \\
 & Penultimate & 60 & $+0.154$ & $+0.114$ & ns \\
 & Logit       & 48 & $+0.122$ & $+0.086$ & ns \\
\bottomrule
\end{tabular}
\end{table}

\begin{table}[!htbp]
\caption{\textbf{Per-experiment $\delta$(RDM) correlations at the penultimate layer.}
For each experiment and RD signature, Spearman $r$ is computed between the logistic nonlinearity index $k_{\mathrm{fit}}$ and the signature mean across the 15 models (raw values; $n = 15$ per experiment). Partial correlations (controlling for accuracy) are not reported per experiment due to insufficient degrees of freedom ($n = 15$); pooled partial correlations are reported in Supplementary Table~8.}
\label{tab:supp_delta_rdm_experiments}
\centering
\small
\setlength{\tabcolsep}{5pt}
\renewcommand{\arraystretch}{1.2}
\begin{tabular}{lrrr}
\toprule
Experiment & AUC ($r$) & $\kappa$ ($r$) & $\beta$ ($r$) \\
\midrule
Contrast         & $-0.036$ & $+0.346$ & $-0.146$ \\
EidolonI         & $-0.154$ & $+0.093$ & $+0.132$ \\
Phase scrambling & $+0.371$ & $-0.261$ & $+0.393$ \\
Uniform noise    & $+0.500$ & $-0.396$ & $+0.436$ \\
\midrule
\textit{Overall (pooled)} & $-0.069\ /\ +0.355^{*}$ &
$+0.154\ /\ +0.114^{\mathrm{ns}}$ &
$-0.259\ /\ -0.312^{\mathrm{ns}}$ \\
\bottomrule
\end{tabular}
\par\smallskip
\begin{minipage}{\linewidth}
\footnotesize\textbf{Notes.} Overall row shows raw $r$ / partial $r$ (accuracy-controlled) pooled across all four experiments ($n = 60$). The sign reversal in AUC from raw ($-0.069$) to partial ($+0.355$) reflects masking by the shared accuracy signal.
\end{minipage}
\end{table}

\FloatBarrier
\FloatBarrier

\subsection*{Supplementary Note 7: Full RD statistics of the untrained ResNet-50 baseline}
\label{sec:supp_untrained}

We report complete RD statistics for the ten randomly initialized ResNet-50 seeds alongside all trained ModelZoo families. Signatures are aggregated to the source level (median across 12 distortion experiments per seed or model) before computing family-level summaries. AUC and $\kappa$ are computed on the experiment-normalized cost axis (cost rescaled to $[0,1]$ within each experiment, enabling cross-experiment comparability). $\beta$ is reported on the raw cost axis, where untrained models produce numerically stable estimates. Kruskal-Wallis tests are conducted on source-level aggregated values across all families including humans and untrained models.

\textbf{Untrained model behavior.} Across all 850 confusion matrices (10 seeds $\times$ 85 experiment$\times$condition combinations), untrained models achieve $6.2\% \pm 0.2\%$ accuracy, consistent with 16-way chance. Eight of ten seeds produce confusion matrices that collapse entirely to a single predicted category, yielding $\beta \approx 0$ (no rate--distortion tradeoff) and $\kappa \gg 10^5$ (degenerate frontier geometry). Two seeds (391, 967) split responses across two categories, yielding partial frontier structure ($\beta \approx -1.38$). This bimodality reflects random initialization variance rather than any meaningful compression strategy. All seeds are reported for completeness.

\textbf{Cross-family comparisons.} The Kruskal-Wallis omnibus test across all family groups (source-aggregated medians; $n = 1$ for Humans, $n = 2$--$4$ per trained family, $n = 10$ for Untrained) is significant for $\kappa$ ($H = 20.74$, $p = 0.004$) and accuracy ($H = 22.16$, $p = 0.002$), but not for AUC ($H = 10.28$, $p = 0.17$) or $\beta$ ($H = 13.94$, $p = 0.052$). The non-significance of AUC reflects that the untrained AUC (mean $= 0.29 \pm 0.13$) is not systematically different from trained models on this axis, since the wide seed-to-seed variance traces to which random category receives collapsed predictions, not to any learned structure. The marginal non-significance of $\beta$ reflects within-group bimodality (8 seeds near zero, 2 near $-1.38$). The near-zero $\beta$ of the eight collapsed seeds is itself the interpretively meaningful signal, marking the complete absence of a compression tradeoff and qualitatively distinct from all trained models.

\begin{table}[!htbp]
\caption{\textbf{RD signature descriptives by model family, including untrained baseline.}
Source-level medians (median across 12 experiments per source) aggregated within each family. $n$: number of unique sources (seeds for Untrained). AUC and $\log_{10}(\kappa)$ use the experiment-normalized cost axis; $\beta$ uses the raw cost axis.}
\label{tab:supp_untrained}
\centering
\small
\setlength{\tabcolsep}{5pt}
\renewcommand{\arraystretch}{1.2}
\begin{tabular}{lrcccc}
\toprule
Family & $n$ & Accuracy & AUC & $\log_{10}(\kappa)$ & $\beta$ \\
\midrule
Humans             & 1  & 0.701 & 0.283 & 1.86 & $-0.245$ \\
VisionTransformers & 3  & 0.894 & 0.322 & 1.81 & $-0.185$ \\
LargeCNNs          & 2  & 0.773 & 0.292 & 2.02 & $-0.237$ \\
SemiSupervised     & 2  & 0.803 & 0.316 & 2.43 & $-0.249$ \\
SelfSupervised     & 4  & 0.616 & 0.273 & 2.50 & $-0.277$ \\
ShapeBiased        & 2  & 0.653 & 0.274 & 2.67 & $-0.373$ \\
LocalModels        & 2  & 0.203 & 0.216 & 4.07 & $-1.415$ \\
\midrule
Untrained (mean $\pm$ SD) & 10 & $0.062 \pm 0.002$ & $0.293 \pm 0.125$ & $5.65 \pm 1.04$ & $-0.277 \pm 0.583$ \\
\bottomrule
\end{tabular}
\par\smallskip
\begin{minipage}{\linewidth}
\footnotesize\textbf{Notes.} Untrained $\beta$ mean and SD are dominated by the bimodal seed distribution (8 seeds: $\beta \approx 0$; 2 seeds: $\beta \approx -1.38$); the median across untrained seeds is $\approx 0$. Kruskal-Wallis omnibus: $\kappa$, $H = 20.74$, $p = 0.004$; accuracy, $H = 22.16$, $p = 0.002$; AUC, $H = 10.28$, $p = 0.17$; $\beta$, $H = 13.94$, $p = 0.052$.
\end{minipage}
\end{table}

\FloatBarrier

\subsection*{Supplementary Note 8: Full statistical results for the degradation nonlinearity analysis}
\label{sec:supp_degradation}

The nonlinearity index was computed from accuracy trajectories across graded severity levels for each system--experiment pair ($n = 285$; 41 systems $\times$ 11 experiments with at least four severity levels and a monotone mean accuracy gradient). Experiments excluded for lacking monotone severity gradients (rotation, colour, false-colour, power-equalisation) and the noise-experiment (humans only) are not included. The nonlinearity index is defined as the RMSE of accuracy from a linear fit across ordered severity levels,
where higher values indicate more abrupt, threshold-like accuracy collapse. All signatures were aggregated as median values across severity conditions within each system--experiment pair prior to correlation. Spearman rank correlations were computed pooled across all system--experiment pairs. Linear mixed models included experiment as a random intercept and mean accuracy as a covariate, and predictors were z-scored within each corpus prior to fitting. Within-experiment demeaned analyses removed all between-experiment variance by
subtracting experiment-level means from both predictors and the outcome before fitting ordinary least-square regressions. The experiment-type moderation distinguished eight noise-corruption perturbations (uniform noise, noise-png, salt-and-pepper-png, highpass, lowpass, EidolonI--III) from three signal-manipulation perturbations (contrast, contrast-png, phase scrambling) and was implemented as a $\kappa \times \mathrm{exp\_type}$ interaction term in the linear mixed model. BH--FDR correction was applied across all signature--analysis combinations. Corpus-split analyses were run independently on the GEN ($n = 157$; 26 models) and ModelZoo ($n = 128$; 16 models) subsets with within-corpus z-scoring.

\textbf{AUC.} The negative AUC--nonlinearity relationship was consistent across analyses and experiment types. The pooled Spearman correlation was $r = -0.413$ ($n = 285$). In a linear mixed model controlling for accuracy, AUC was a significant predictor of smoother degradation (coefficient $= -0.039$,
$\mathrm{SE} = 0.014$, $t = -2.78$, $p = 0.006$). The within-experiment demeaned analysis confirmed this was not an artifact of between-experiment confounds ($t = -5.66$, $p = 3.7 \times 10^{-8}$). In the noise-corruption-only subset, the AUC effect was particularly strong ($r = -0.518$; mixed model
coefficient $= -0.118$, $t = -10.44$, $p < 0.0001$; $\Delta$pseudo-$R^2 = 0.327$). The experiment-type moderation for AUC was significant (LRT $\chi^2(1) = 34.97$, $p < 0.0001$) but reflects a range artifact where AUC is near-perfectly correlated with accuracy in both experiment types ($r_{\mathrm{AUC,acc}} = 0.959$ signal, $0.915$ noise), and the interaction disappears in direction after correcting for this (AUC residualized on accuracy: noise-corruption slope $= -0.046$, $p < 0.0001$; signal-manipulation slope $= +0.032$, $p = 0.003$; interaction LRT $p < 0.0001$). The AUC sign reversal across experiment types should therefore not be interpreted as a genuine reversal of the AUC--smoothness relationship but as an accuracy-range artifact; the within-experiment demeaned result is the appropriate summary. Corpus-split analyses confirmed the AUC effect in both GEN ($r = -0.401$; LRT $p < 0.0001$) and ModelZoo ($r = -0.424$; LRT $p = 0.018$) independently.

\textbf{$\kappa$.} The pooled Spearman correlation between $\log_{10}(\kappa + 1)$ and degradation nonlinearity was $r = +0.381$ ($n = 285$). In a linear mixed model controlling for accuracy, $\kappa$ was a significant predictor ($t = 2.14$, $p = 0.033$; $\Delta$pseudo-$R^2 = 0.012$). However, the within-experiment demeaned analysis showed only a marginal effect ($t = 1.85$, $p = 0.066$), indicating partial dependence on between-experiment variation in $\kappa$ range. The experiment-type moderation was highly significant (LRT
$\chi^2(1) = 36.17$, $p < 0.0001$), with $\kappa$ predicting more abrupt degradation within noise-corruption experiments ($\beta_{\mathrm{within}} = +0.039$, $\mathrm{SE} = 0.008$, $p < 0.0001$) and less abrupt degradation within signal-manipulation experiments ($\beta_{\mathrm{within}} = -0.054$, $\mathrm{SE} = 0.014$, $p = 0.0002$). This sign reversal was consistent across individual experiments (9/11 in the predicted direction; binomial $p = 0.033$) and replicated independently in both the GEN corpus ($\beta_{\mathrm{int}} = 0.067$, $p < 0.0001$) and ModelZoo corpus ($\beta_{\mathrm{int}} = 0.048$, $p < 0.0001$). A sensitivity analysis reclassifying phase scrambling as noise-corruption (leaving only contrast as signal-manipulation) attenuated but did not eliminate the moderation (LRT $p = 0.011$). The noise-corruption-only Spearman correlation was $r = +0.468$
($n = 199$; mixed model $t = 4.70$, $p < 0.0001$; $\Delta$pseudo-$R^2 = 0.092$).

\textbf{$\beta$.} $\beta$ did not independently predict degradation nonlinearity in any analysis after accuracy control. The pooled Spearman correlation was $r = -0.339$ ($n = 285$), but this reflects shared dependence on accuracy ($r_{\beta,\mathrm{acc}} = -0.259$). In the linear mixed model controlling for accuracy, $\beta$ was not significant ($t = -1.12$, $p = 0.26$; $\Delta$pseudo-$R^2 = 0.001$). The within-experiment demeaned analysis confirmed no independent effect ($t = 0.57$, $p = 0.57$). However, $\beta$ did show a significant experiment-type moderation (LRT $\chi^2(1) = 24.22$, $p < 0.0001$), with a positive slope in signal-manipulation experiments ($\beta_{\mathrm{within}} = +0.037$, $p = 0.0009$) and a near-zero and non-significant slope in noise-corruption experiments ($\beta_{\mathrm{within}} = -0.009$, $p = 0.150$). This moderation is asymmetric --- $\beta$ matters for signal experiments but not noise experiments --- and does not reflect an independent signal beyond accuracy given the null within-experiment demeaned result.

\begin{table}[!htbp]
\caption{\textbf{Degradation nonlinearity analysis: full
statistical results for all three RD signatures.}
Each row reports one analysis level for one signature. Mixed models include experiment as a random intercept and z-scored mean accuracy as a covariate. All predictors are z-scored. Within-experiment demeaned analyses use OLS on
experiment-mean-subtracted variables. The moderation interaction term is $\mathrm{signature} \times \mathrm{exp\_type}$ in the mixed model. Corpus-split values report the $\kappa \times \mathrm{exp\_type}$ interaction coefficient for GEN and ModelZoo independently. $n_{\mathrm{total}} = 285$; $n_{\mathrm{noise}} = 199$; $n_{\mathrm{signal}} = 86$. BH--FDR correction applied within each signature across analysis levels.}
\label{tab:supp_degradation}
\centering
\small
\setlength{\tabcolsep}{4pt}
\renewcommand{\arraystretch}{1.25}
\begin{tabular}{llrrrrl}
\toprule
Signature & Analysis & $n$ & Stat. & Value & $p$ & Sig. \\
\midrule
\multirow{9}{*}{AUC}
 & Pooled Spearman $r$ & 285 & $r$ & $-0.413$ & $<0.0001$ & *** \\
 & Mixed model (acc.\ controlled) & 285 & $t$ & $-2.78$ & $0.006$ & ** \\
 & Within-exp.\ demeaned & 285 & $t$ & $-5.66$ & $3.7\times10^{-8}$ & *** \\
 & Noise-only Spearman $r$ & 199 & $r$ & $-0.518$ & $<0.0001$ & *** \\
 & Noise-only mixed model & 199 & $t$ & $-10.44$ & $<0.0001$ & *** \\
 & Noise-only $\Delta R^2$ & 199 & $\Delta R^2$ & $0.327$ & --- & \\
 & Exp.-type moderation LRT & 285 & $\chi^2(1)$ & $34.97$ & $<0.0001$ & *** \\
 & Corpus split: GEN & 157 & LRT $p$ & --- & $<0.0001$ & *** \\
 & Corpus split: ModelZoo & 128 & LRT $p$ & --- & $0.018$ & * \\
\midrule
\multirow{11}{*}{$\kappa$}
 & Pooled Spearman $r$ & 285 & $r$ & $+0.381$ & $<0.0001$ & *** \\
 & Mixed model (acc.\ controlled) & 285 & $t$ & $+2.14$ & $0.033$ & * \\
 & $\Delta R^2$ (mixed model) & 285 & $\Delta R^2$ & $0.012$ & --- & \\
 & Within-exp.\ demeaned & 285 & $t$ & $+1.85$ & $0.066$ & ns \\
 & Noise-only Spearman $r$ & 199 & $r$ & $+0.468$ & $<0.0001$ & *** \\
 & Noise-only mixed model & 199 & $t$ & $+4.70$ & $<0.0001$ & *** \\
 & Noise-only $\Delta R^2$ & 199 & $\Delta R^2$ & $0.092$ & --- & \\
 & Exp.-type moderation LRT & 285 & $\chi^2(1)$ & $36.17$ & $<0.0001$ & *** \\
 & Within noise-corruption & 199 & $\beta$ & $+0.039$ & $<0.0001$ & *** \\
 & Within signal-manipulation & 86 & $\beta$ & $-0.054$ & $0.0002$ & *** \\
 & Corpus split: GEN & 157 & $\beta_{\mathrm{int}}$ & $+0.067$ & $<0.0001$ & *** \\
 & Corpus split: ModelZoo & 128 & $\beta_{\mathrm{int}}$ & $+0.048$ & $<0.0001$ & *** \\
 & Binomial sign consistency & 11 & $9/11$ & --- & $0.033$ & * \\
 & Sensitivity (phase as noise) & 285 & LRT $p$ & --- & $0.011$ & * \\
\midrule
\multirow{6}{*}{$\beta$}
 & Pooled Spearman $r$ & 285 & $r$ & $-0.339$ & $<0.0001$ & *** \\
 & Mixed model (acc.\ controlled) & 285 & $t$ & $-1.12$ & $0.26$ & ns \\
 & $\Delta R^2$ (mixed model) & 285 & $\Delta R^2$ & $0.001$ & --- & \\
 & Within-exp.\ demeaned & 285 & $t$ & $+0.57$ & $0.57$ & ns \\
 & Exp.-type moderation LRT & 285 & $\chi^2(1)$ & $24.22$ & $<0.0001$ & *** \\
 & Within noise-corruption & 199 & $\beta$ & $-0.009$ & $0.150$ & ns \\
 & Within signal-manipulation & 86 & $\beta$ & $+0.037$ & $0.0009$ & *** \\
\bottomrule
\end{tabular}
\par\smallskip
\begin{minipage}{\linewidth}
\footnotesize\textbf{Notes.} Mixed model: $\mathrm{RMSE} \sim \mathrm{acc\_z} + \mathrm{sig\_z} + (1|\mathrm{experiment})$, ML estimation. Within-experiment demeaned: OLS on experiment-mean-subtracted variables. $\Delta R^2$: pseudo-$R^2$ increment over accuracy-only baseline (residual variance reduction). Moderation: $\mathrm{RMSE} \sim \mathrm{acc\_z} + \mathrm{sig\_z} \times \mathrm{exp\_type} + (1|\mathrm{experiment})$. Stars: $^{***}p < 0.001$, $^{**}p < 0.01$, $^{*}p < 0.05$, ns not significant. AUC experiment-type moderation reflects an accuracy-range artifact (see Note text), thus the within-experiment demeaned result is the appropriate summary for AUC.
\end{minipage}
\end{table}

\begin{figure}[!htbp]
  \centering
  \includegraphics[width=\textwidth]{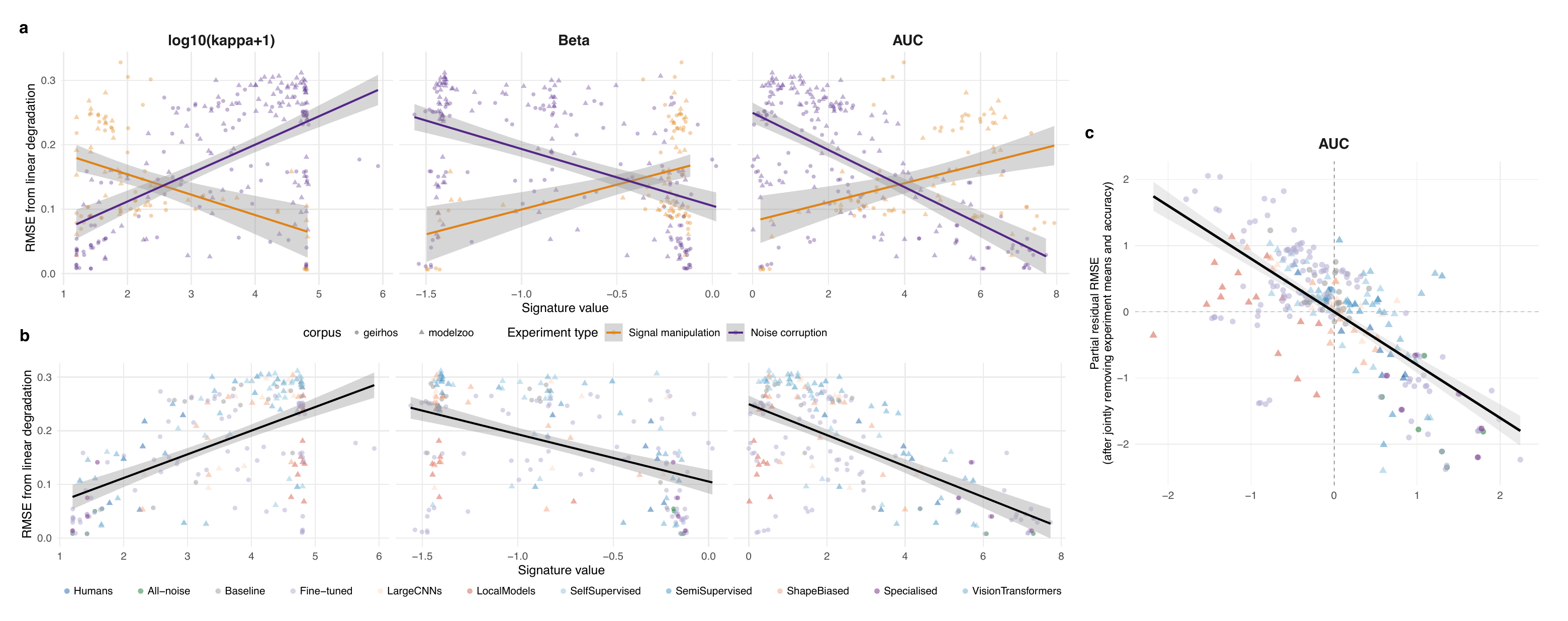}
    \caption{\textbf{Degradation nonlinearity analysis: scatter plots for all three RD signatures.}
    \textbf{(a)} All three signatures plotted against degradation nonlinearity (RMSE from linear accuracy trajectory), coloured by experiment type (noise-corruption: purple; signal-manipulation: orange). Separate regression lines per experiment type illustrate the sign reversal in $\kappa$ and $\beta$ and the consistent negative direction of AUC.
    \textbf{(b)} Noise-corruption experiments only ($n = 199$), coloured by system family. The positive $\kappa$--nonlinearity and negative AUC--nonlinearity relationships are visible across families.
    \textbf{(c)} Within-experiment demeaned scatter for AUC: both AUC and RMSE are residualized by subtracting experiment-level means, removing all between-experiment confounding. The negative slope ($t = -5.66$, $p = 3.7 \times 10^{-8}$) confirms that AUC independently predicts smoother degradation within
    experiments.}
    \label{fig:supp_degradation}
\end{figure}

\subsection*{Supplementary Note 9: Compute Resources and Data Availability}
\label{app:compute_licenses}

\paragraph{Computational resources.}
Primary RD inference (MAP optimization and Blahut--Arimoto) is CPU-based and was parallelized across an HPC cluster. The pipeline has three steps with distinct cost profiles. Step~1 - MAP inference of the latent distortion matrix $\rho$ - is the computational bottleneck, scaling as $O(K^2)$ in both parameter space and the inner Blahut—Arimoto loop. For $K=16$ categories, MAP inference takes approximately 50 minutes per confusion matrix on 2 CPU cores, the full empirical analysis covers 1,650 confusion matrices (all models and conditions in the GEN benchmark repository), corresponding to approximately 1,375 core-hours in total. All jobs were parallelized across an HPC cluster, significantly reducing wall-clock time. Steps~2 and~3 - RD frontier tracing and signature extraction ($\beta$, $\kappa$, AUC) - run in seconds given $\rho$ and add negligible overhead. 
Activation centroid extraction for the layer-resolved internal representations analysis required GPU computation (NVIDIA A100 nodes on the Minerva HPC cluster at Mount Sinai). Each of the 15 models was processed in a separate array job, requiring approximately 30--90 minutes per model across 31 conditions and 5 layer depths.

\paragraph{Data and model licenses.}
The GEN benchmark~\citep{geirhos2018generalisation} and ModelZoo repository~\citep{geirhos2021partial} are both released under the MIT License and are publicly available at \url{https://github.com/bethgelab/model-vs-human}. The human psychophysics data from both repositories is included in the same repository under the same license terms. All pre-trained CNN models (GoogLeNet, ResNet-152, VGG-19, and the robustness-trained variants) were accessed via ModelZoo and are subject to their respective original licenses. Each is cited with its original paper in the main text. No new datasets or models are introduced in this work.


\end{document}